\pdfoutput=1

\documentclass[11pt]{article}

\usepackage{ACL2023}

\usepackage{times}
\usepackage{latexsym}

\usepackage[T1]{fontenc}

\usepackage[utf8]{inputenc}

\usepackage{microtype}

\usepackage{inconsolata}

\usepackage{xcolor} %
\usepackage{soul} %

\usepackage{multirow} %
\usepackage{booktabs}
\usepackage{adjustbox}

\usepackage{linguex}
\usepackage{relsize}

\usepackage{caption}
\usepackage{subcaption}

\usepackage{todonotes}

\usepackage[]{titlesec}
\titlespacing*{\paragraph}{0pt}{1ex}{1ex}
\usepackage{enumitem}

\makeatletter
\newcommand*\iftodonotes{\if@todonotes@disabled\expandafter\@secondoftwo\else\expandafter\@firstoftwo\fi}  %
\makeatother

\definecolor{controlshuf}{HTML}{606060}
\definecolor{nondetshuf}{HTML}{E8384F}
\definecolor{detshuf21}{HTML}{FFB000}
\definecolor{detshuf57}{HTML}{8db000}
\definecolor{detshuf84}{HTML}{62BB35}
\definecolor{locshuf3}{HTML}{208EA3}
\definecolor{locshuf5}{HTML}{4178BC}
\definecolor{locshuf10}{HTML}{AA71FF}
\definecolor{evenodd}{HTML}{E37CFF}

\definecolor{controlrev}{HTML}{606060}
\definecolor{partialrev}{HTML}{E5A836}
\definecolor{fullrev}{HTML}{A348A6}

\definecolor{NoHop}{HTML}{606060}
\definecolor{tokenhop}{HTML}{fa8128}
\definecolor{wordhop}{HTML}{03a0ff}

\newcommand{\singularmarker}{%
  \setlength{\fboxsep}{1pt}%
  \fbox{\texttt{S}}%
}
\newcommand{\pluralmarker}{%
  \setlength{\fboxsep}{1pt}%
  \fbox{\texttt{P}}%
}
\newcommand{\revmarker}{%
  \setlength{\fboxsep}{1pt}%
  \fbox{\texttt{R}}%
}

\newcommand{\highlight}[2]{%
    \begingroup
    \definecolor{hlcolor}{HTML}{#1}%
    \sethlcolor{hlcolor}%
    \hl{#2}%
    \endgroup
}

\newcommand{\tokenHe}{%
  \highlight{FDFD95}{He}%
}
\newcommand{\tokenThey}{%
  \highlight{FDFD95}{They}%
}
\newcommand{\tokencleans}{%
  \highlight{FDE4CF}{ cleans}%
}
\newcommand{\tokenclean}{%
  \highlight{FDE4CF}{ clean}%
}
\newcommand{\tokenhis}{%
  \highlight{FFCFD2}{ his}%
}
\newcommand{\tokenvery}{%
  \highlight{F1C0E8}{ very}%
}
\newcommand{\tokenmessy}{%
  \highlight{CFBAF0}{ messy}%
}
\newcommand{\tokenbooks}{%
  \highlight{A3C4F3}{ books}%
}
\newcommand{\tokenhe}{%
  \highlight{90DBF4}{he}%
}
\newcommand{\tokenlf}{%
  \highlight{98F5E1}{lf}%
}
\newcommand{\tokenperiod}{%
  \highlight{B9FBC0}{.}%
}
\newcommand{\tokensing}{%
  \highlight{FF77E6}{\singularmarker}%
}
\newcommand{\tokenplur}{%
  \highlight{3EC8FF}{\pluralmarker}%
}
\newcommand{\tokenrev}{%
  \highlight{8DE969}{\revmarker}%
}

\title{
Mission: Impossible Language Models
}

\author{Julie Kallini$^{1}$,
  Isabel Papadimitriou$^{1}$,
  Richard Futrell$^{2}$, \\
  \textbf{Kyle Mahowald}$^{3}$\textbf{,}
  \textbf{Christopher Potts}$^{1}$ \\[1ex]
  ${}^{1}$Stanford University; ${}^{2}$University of California, Irvine; ${}^{3}$University of Texas, Austin \\[1ex]
  \texttt{kallini@stanford.edu}
}

\usepackage[nameinlink,capitalize]{cleveref}

\begin{document}

\setlength{\Exlabelwidth}{1em}
\setlength{\Exlabelsep}{0.7em}
\setlength{\SubExleftmargin}{1.3em}
\setlength{\Extopsep}{2pt}

\maketitle

\begin{abstract}
Chomsky and others have very directly claimed that large language models (LLMs) are equally capable of learning languages that are possible and impossible for humans to learn. However, there is very little published experimental evidence to support such a claim. Here, we develop a set of synthetic \emph{impossible languages} of differing complexity, each designed by systematically altering English data with unnatural word orders and grammar rules. These languages lie on an impossibility continuum: at one end are languages that are inherently impossible, such as random and irreversible shuffles of English words, and on the other, languages that may not be intuitively impossible but are often considered so in linguistics, particularly those with rules based on counting word positions. We report on a wide range of evaluations to assess the capacity of GPT-2 small models to learn these uncontroversially impossible languages, and crucially, we perform these assessments at various stages throughout training to compare the learning process for each language. Our core finding is that GPT-2 struggles to learn impossible languages when compared to English as a control, challenging the core claim. More importantly, we hope our approach opens up a productive line of inquiry in which different LLM architectures are tested on a variety of impossible languages in an effort to learn more about how LLMs can be used as tools for these cognitive and typological investigations.
\end{abstract}

\section{Introduction}

\begin{figure}[ht]
    \centering
    \includegraphics[width=0.45\textwidth]{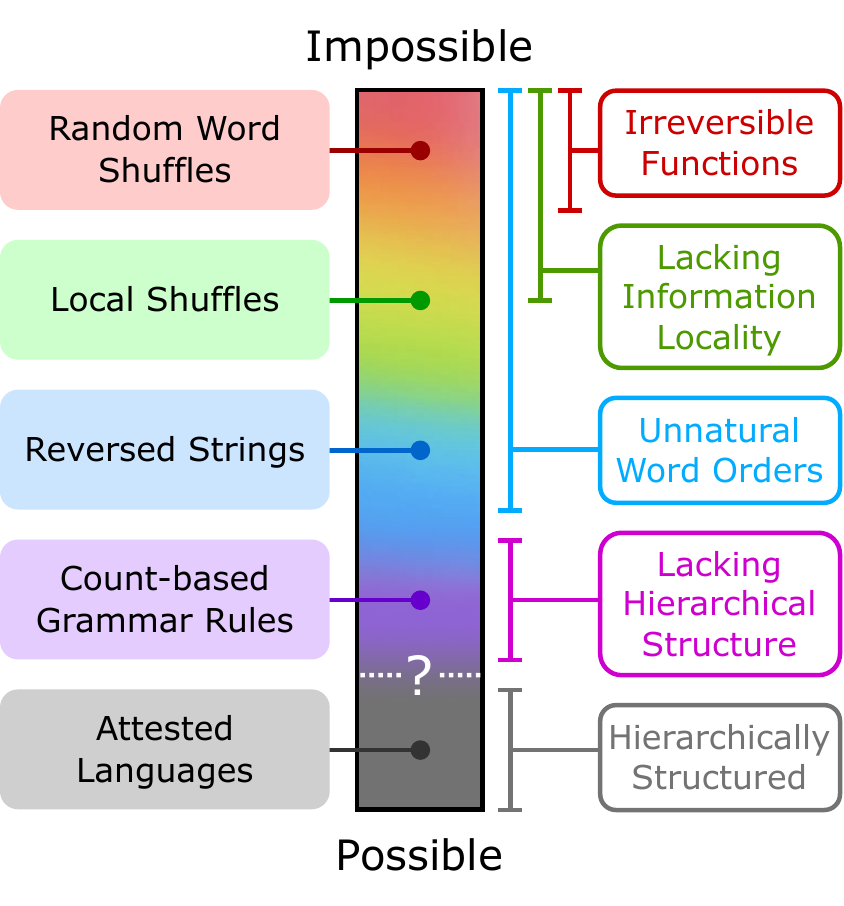}
    \caption{Partial impossibility continuum of languages based on complexity. We assess the learnability of languages at different points in the continuum and push the (currently unclear) boundary between possible and impossible.}
    \label{fig:language-continuum}
\end{figure}

\citet{chomsky2023cowen}, \citet{chomsky2023nyt}, \citet{moro2023impossible}, and \citet{Bolhuis2024} make very broad claims to the effect that large language models (LLMs) are equally capable of learning possible and impossible human languages. For these authors, it follows from this claim that LLMs cannot teach us anything about language, and so the claim (if true) would have significant consequences for linguistic methodology and potentially also for the viability of LLMs as the basis for robust language capabilities.

These authors state this claim in absolute terms. For example, \citet{chomsky2023nyt} flatly assert that LLMs ``are incapable of distinguishing the possible from the impossible,'' \citet{chomsky2023cowen} says this property ``can't be modified,'' and \citet{moro2023impossible} write that ``the distinction between possible versus impossible languages cannot be formulated by definition for LLM.'' \citet{Bolhuis2024} go so far as to claim that ``LLMs can produce ‘impossible’ languages [...] just as well as (if not better than) natural language output.''
One might expect such strong claims to be supported by extensive formal analysis and/or experimental evidence. However, as far as we are aware, this is not the case. The sole experimental paper cited by the above authors is  \citealt{mitchell-bowers-2020-priorless}---an important and inspiring paper but not one that can resolve these questions on its own. In addition, linguists themselves do not even have an agreed upon notion of what defines the possible or the impossible languages, to say nothing of having formal results with respect to LLMs.

Here we provide extensive new experimental evidence to inform the claim that LLMs are equally capable of learning possible and impossible languages in the human sense. Arguably, the central challenge for such work is the fact that there is no agreed-upon way of distinguishing these two groups. We do not feel positioned ourselves to assert such a definition, so we instead offer some examples of impossible languages on a continuum of intuitive complexity (\Cref{fig:language-continuum}). 
Some of these examples seem intuitively impossible, such as random sentence-level shuffling of English words. Others operationalize less obvious but common claims in the linguistics literature about rules that are impossible, like those that depend on counting words. 

All of our examples are, we take it, uncontroversial instances of impossible languages. Thus, our experiments can inform the core hypotheses as follows: if LLMs learn these languages as well as they learn natural languages, then the claims of Chomsky and others are supported (for the specific class of LLMs tested). Conversely, if LLMs do not learn these languages as well as the possible ones, it would call into question those assertions. In that case, proponents of those claims ought to provide examples of impossible languages that they find more informative, which we can then evaluate using our approach to further advance the discussion.

Our experiments use GPT-2 small models \cite{radford2018improving, radford2019languagemodels},
and our base training corpus is the BabyLM dataset \cite{warstadt2023papers}, which we modify in various ways to implement our impossible languages.
What we find is that these models indeed struggle to learn impossible languages, shown through three core experiments:

\begin{itemize}[topsep=2pt, itemsep=0pt, leftmargin=12pt]
    \item In \textbf{Experiment 1}, we train GPT-2 models on our set of defined possible and impossible languages, measuring their learning efficiency through test set perplexities. We find that \emph{models trained on possible languages learn more efficiently}, evident from lower perplexities achieved in fewer training steps.
    \item In \textbf{Experiment 2}, we more closely examine a set of languages that exhibit count-based verb marking rules, using surprisal comparisons to target the relevant patterns. We find that \mbox{GPT-2s} trained on possible languages are more surprised by ungrammatical constructions, indicating that \emph{models disprefer agreement rules involving counting}.
    \item In \textbf{Experiment 3}, we dive deeper into the internal mechanisms that models may develop to learn such count-based grammar rules using causal abstraction analysis. We find that \emph{models develop natural, modular solutions to unnatural grammatical patterns}.
\end{itemize}

Overall, our experimental results strongly challenge the claims of Chomsky and others given above, and we believe they pave the way for even deeper discussions of LLMs as models of language learning. At the same time, we recognize that models and humans exhibit fundamental differences, but the extent to which models favor or disfavor natural languages can be influenced by specific architectural decisions (as demonstrated by our findings on tokenization and positional encodings). We hope this paper initiates a new line of work that explores how different model architectures can distinguish between the possible and impossible languages.\footnote{The code for this paper is available at \url{https://github.com/jkallini/mission-impossible-language-models}.}

\section{Background and Related Work}

\subsection{Impossible Human Languages and Language Universals}

The notion of an impossible human language is elusive and difficult to define, in part due to a lack of consensus on which properties are universal in human language and which properties are ``impossible'' \citep{comrie1989language, evans2009myth, Nefdt_2024}.
For instance, \emph{recursion}, or the principle that all languages produce hierarchical syntactic structures via recursive procedures, has been claimed to be a universal property of human language \citep{chomsky1957syntactic, chomsky1965aspects, chomsky2002nature, hauser2002faculty}.
However, the motivations for recursion have been questioned, with empirical limits on the maximum depth of nested phrases \cite{karlsson2007constraints, jin-etal-2018-depth} and counterevidence from at least one natural language that seems to lack embedded structures \cite{everett2012piraha}. Still, if we grant that possible languages are defined by hierarchical, recursive rules, what defines the impossible languages? \citet{moro2023impossible} claim that the class of impossible languages would use the ``opposite'' type of rules: those based on the linear order of words. \citet{musso2003broca} provide a few concrete examples that involve counting word positions to mark features like negation and agreement, and we include languages with similar rules in our set of tested impossible languages.

It is important to also distinguish what is impossible from what is merely typologically marked, such as the word order patterns listed in  \citeposs{greenberg1963universals} language universals. Previous work has shown that such word order universals can arise through a language's optimization of communication efficiency, achieved by balancing complexity and ambiguity \cite{hahnetal2020universals, futrell2022information}.
While our current exploration does not encompass attested languages, various impossible languages can similarly differ in their information-theoretic complexity, informing the patterns that lie at the boundary between possible and impossible.

\subsection{Training Language Models with Unnatural Word Orders}

The only work cited by Chomsky that investigates neural language models' ability to learn impossible languages is \citealt{mitchell-bowers-2020-priorless}, which finds that recurrent neural networks (RNNs; \citealp{elman1990finding}) trained on various unnatural language constructs, such as reversed sentences and randomized vocabularies, achieve high accuracy on a subject--verb number agreement task.
Other work turns to more recent Transformer-based language models \cite{vaswani2017attention}, observing their sensitivity to word order and phrase structure \cite{alleman-etal-2021-syntactic, galke2023makes} as well as their surprising ability to learn from syntactic information alone \cite{huang2023lexinvariant}.
Studies by \citet{sinha-etal-2021-masked} and \citet{abdou-etal-2022-word} debate the impact of tokenization, pretraining adjustments, and positional encodings in recovering word order information from shuffled languages.
Further investigations into BERT's \cite{devlin-etal-2019-bert} reliance on word order for grammatical role classification suggest that lexical cues alone may not always be sufficient for good performance (\citealp{papadimitriou-etal-2022-classifying-grammatical}; see also \citealp{hessel-schofield-2021-effective, pham-etal-2021-order}).

\subsection{Language Models and Formal Languages}

A related line of research examines the abilities of neural language models to express formal languages, as defined by the \emph{Chomsky hierarchy}
\cite{chomsky1956models,chomsky1959certain}. Human language is considered to be slightly more expressive than context-free languages due to certain syntactic phenomena that interleave constituents \cite{shieber1985evidence,joshi1985TAG}.
Previous work has shown that RNNs or related models can represent variants of counter and \textsc{Dyck} languages, which are context-free \cite{weiss-etal-2018-practical, merrill-2019-sequential, merrill-etal-2020-formal, hewitt-etal-2020-rnns}.\footnote{Though counter and \textsc{Dyck} languages are context-free, some of the variants in the cited work are regular.}
Similar work on Transformer architectures has shown that, while they are theoretically Turing-complete provided arbitrary precision and decoder steps \cite{perez2021attention}, they cannot empirically model many regular and non-regular languages \cite{hahn2020theoretical, ebrahimi-etal-2020-self, deletang2023neural}.

The inability of Transformer-based language models to learn more complex languages in the Chomsky hierarchy seems surprising, given their impressive performance on natural language. This could be interpreted as evidence that theoretically weak computational models are sufficient for expressing human language. Alternatively, Transformer-based models can be augmented to have inductive biases for nested, hierarchical structures through architecture changes, like the addition of a stack component  \cite{hao-etal-2018-context, murty2023pushdown}, or data-centered approaches, like structural pretraining \cite{papadimitriou2023injecting}.

\section{Impossible Languages} \label{sec:impossible-languages}

\begin{table*}
  \centering
  \resizebox{1.0\textwidth}{!}{%
  \begin{tabular}{l|l|l|l}
    \toprule
    \textbf{Class} & \textbf{Language} & \textbf{Example 1} & \textbf{Example 2} \\ \bottomrule
    & & & \\[-1em]
    \multirow{12}{*}{\textsc{*Shuffle}} & \textcolor{controlshuf}{\textsc{NoShuffle}}              & 
    \texttt{\tokenHe\,\tokencleans\,\tokenhis\,\tokenvery\,\tokenmessy\,\tokenbooks\,\tokenhe\,\tokenlf\,\tokenperiod} & \texttt{\tokenThey\,\tokenclean\,\tokenhis\,\tokenvery\,\tokenmessy\,\tokenbooks\,\tokenhe\,\tokenlf\,\tokenperiod} \\[3pt] \cline{2-4} 
    & & & \\[-1em]
    & \textcolor{nondetshuf}{\textsc{NondeterministicShuffle}}        & \texttt{\tokenmessy\,\tokenbooks\,\tokenhis\,\tokenhe\,\tokenvery\,\tokenperiod\,\tokenlf\,\tokenHe\,\tokencleans} & \texttt{\tokenhis\,\tokenperiod\,\tokenvery\,\tokenhe\,\tokenThey\,\tokenmessy\,\tokenlf\,\tokenbooks\,\tokenclean} \\[3pt] \cline{2-4}
    & & & \\[-1em]
    & \textcolor{detshuf21}{\textsc{DeterministicShuffle}$(s=21)$}   & \texttt{\tokencleans\,\tokenHe\,\tokenmessy\,\tokenbooks\,\tokenhe\,\tokenlf\,\tokenvery\,\tokenperiod\,\tokenhis} & \texttt{\tokenclean\,\tokenThey\,\tokenmessy\,\tokenbooks\,\tokenhe\,\tokenlf\,\tokenvery\,\tokenperiod\,\tokenhis} \\[3pt] \cline{2-4}
    & & & \\[-1em]
    & \textcolor{detshuf57}{\textsc{DeterministicShuffle}$(s=57)$}   & \texttt{\tokencleans\,\tokenhis\,\tokenHe\,\tokenmessy\,\tokenperiod\,\tokenhe\,\tokenvery\,\tokenlf\,\tokenbooks} & \texttt{\tokenclean\,\tokenhis\,\tokenThey\,\tokenmessy\,\tokenperiod\,\tokenhe\,\tokenvery\,\tokenlf\,\tokenbooks} \\[3pt] \cline{2-4}
    & & & \\[-1em]
    & \textcolor{detshuf84}{\textsc{DeterministicShuffle}$(s=84)$}   & \texttt{\tokenHe\,\tokenmessy\,\tokenperiod\,\tokenlf\,\tokenhis\,\tokenvery\,\tokenbooks\,\tokencleans\,\tokenhe} & \texttt{\tokenThey\,\tokenmessy\,\tokenperiod\,\tokenlf\,\tokenhis\,\tokenvery\,\tokenbooks\,\tokenclean\,\tokenhe} \\[3pt] \cline{2-4}
    & & & \\[-1em]
    & \textcolor{locshuf3}{\textsc{LocalShuffle}$(w=3)$}            & \texttt{\tokenhis\,\tokenHe\,\tokencleans\,\tokenbooks\,\tokenvery\,\tokenmessy\,\tokenperiod\,\tokenhe\,\tokenlf} & \texttt{\tokenhis\,\tokenThey\,\tokenclean\,\tokenbooks\,\tokenvery\,\tokenmessy\,\tokenperiod\,\tokenhe\,\tokenlf} \\[3pt] \cline{2-4}
    & & & \\[-1em]
    & \textcolor{locshuf5}{\textsc{LocalShuffle}$(w=5)$}            & \texttt{\tokenhis\,\tokenmessy\,\tokenvery\,\tokenHe\,\tokencleans\,\tokenlf\,\tokenbooks\,\tokenhe\,\tokenperiod} & \texttt{\tokenhis\,\tokenmessy\,\tokenvery\,\tokenThey\,\tokenclean\,\tokenlf\,\tokenbooks\,\tokenhe\,\tokenperiod} \\[3pt] \cline{2-4}
    & & & \\[-1em]
    & \textcolor{locshuf10}{\textsc{LocalShuffle}$(w=10)$}           & \texttt{\tokenmessy\,\tokenbooks\,\tokenhis\,\tokenhe\,\tokenvery\,\tokenperiod\,\tokenlf\,\tokenHe\,\tokencleans} & \texttt{\tokenmessy\,\tokenbooks\,\tokenhis\,\tokenhe\,\tokenvery\,\tokenperiod\,\tokenlf\,\tokenThey\,\tokenclean} \\[3pt] \cline{2-4}
    & & & \\[-1em]
    & \textcolor{evenodd}{\textsc{EvenOddShuffle}}                        & \texttt{\tokenHe\,\tokenhis\,\tokenmessy\,\tokenhe\,\tokenperiod\,\tokencleans\,\tokenvery\,\tokenbooks\,\tokenlf} & \texttt{\tokenThey\,\tokenhis\,\tokenmessy\,\tokenhe\,\tokenperiod\,\tokenclean\,\tokenvery\,\tokenbooks\,\tokenlf} \\[3pt]
    \hline
    & & & \\[-1em]
    \multirow{4}{*}{\textsc{*Reverse}} & \textcolor{controlrev}{\textsc{NoReverse}}              & \texttt{\tokenHe\,\tokencleans\,\tokenhis\,\tokenvery\,\tokenmessy\,\tokenbooks\,\tokenrev\,\tokenhe\,\tokenlf\,\tokenperiod} & \texttt{\tokenThey\,\tokenclean\,\tokenhis\,\tokenrev\,\tokenvery\,\tokenmessy\,\tokenbooks\,\tokenhe\,\tokenlf\,\tokenperiod} \\[3pt] \cline{2-4}
    & & & \\[-1em]
    & \textcolor{partialrev}{\textsc{PartialReverse}}                 & \texttt{\tokenHe\,\tokencleans\,\tokenhis\,\tokenvery\,\tokenmessy\,\tokenbooks\,\tokenrev\,\tokenperiod\,\tokenlf\,\tokenhe} & \texttt{\tokenThey\,\tokenclean\,\tokenhis\,\tokenrev\,\tokenperiod\,\tokenlf\,\tokenhe\,\tokenbooks\,\tokenmessy\,\tokenvery} \\[3pt] \cline{2-4}
    & & & \\[-1em]
    & \textcolor{fullrev}{\textsc{FullReverse}}                        & \texttt{\tokenperiod\,\tokenlf\,\tokenhe\,\tokenrev\,\tokenbooks\,\tokenmessy\,\tokenvery\,\tokenhis\,\tokencleans\,\tokenHe} & \texttt{\tokenperiod\,\tokenlf\,\tokenhe\,\tokenbooks\,\tokenmessy\,\tokenvery\,\tokenrev\,\tokenhis\,\tokenclean\,\tokenThey} \\[3pt] \hline
    & & & \\[-1em]
    \multirow{4}{*}{\textsc{*Hop}} & \textcolor{NoHop}{\textsc{NoHop}}               & \texttt{\tokenHe\,\tokenclean\,\tokensing\,\tokenhis\,\tokenvery\,\tokenmessy\,\tokenbooks\,\tokenhe\,\tokenlf\,\tokenperiod} &  \texttt{\tokenThey\,\tokenclean\,\tokenplur\,\tokenhis\,\tokenvery\,\tokenmessy\,\tokenbooks\,\tokenhe\,\tokenlf\,\tokenperiod} \\[3pt] \cline{2-4}
    & & & \\[-1em]
    & \textcolor{tokenhop}{\textsc{TokenHop}}                & \texttt{\tokenHe\,\tokenclean\,\tokenhis\,\tokenvery\,\tokenmessy\,\tokenbooks\,\tokensing\,\tokenhe\,\tokenlf\,\tokenperiod} &  \texttt{\tokenThey\,\tokenclean\,\tokenhis\,\tokenvery\,\tokenmessy\,\tokenbooks\,\tokenplur\,\tokenhe\,\tokenlf\,\tokenperiod} \\[3pt] \cline{2-4}
    & & & \\[-1em]
    & \textcolor{wordhop}{\textsc{WordHop}} & \texttt{\tokenHe\,\tokenclean\,\tokenhis\,\tokenvery\,\tokenmessy\,\tokenbooks\,\tokenhe\,\tokenlf\,\tokensing\,\tokenperiod} &  \texttt{\tokenThey\,\tokenclean\,\tokenhis\,\tokenvery\,\tokenmessy\,\tokenbooks\,\tokenhe\,\tokenlf\,\tokenplur\,\tokenperiod} \\[3pt]
    \toprule
    \end{tabular}
  }
  \caption{List of impossible languages with examples. Control (`\textsc{No*}') languages have patterns that resemble English. Differently colored
  blocks represent different GPT-2 tokens. }
  \label{tab:perturbations}
\end{table*}

Core to our experiments are the set of \emph{impossible languages} we synthesize. In constructing these artificial counterfactual languages, we consider their information-theoretic attributes relevant to machine learning, such as entropy rate, as well as their formal linguistic characteristics, such as adherence to hierarchical grammatical structures. We believe that our choice of languages broadly spans the impossibility continuum hypothesized in \Cref{fig:language-continuum}.

Concretely, we specify impossible languages by defining \emph{perturbation functions} of English sentences. These perturbation functions map English input sentences to sequences of tokens. We categorize our languages into three classes: \textsc{*Shuffle}, \textsc{*Reverse}, and \textsc{*Hop}, defined in the next subsections. Each class has one control language that represents unaltered English, or a pattern that is very similar to English. \Cref{tab:perturbations} provides examples of perturbed sentences in each language.

\subsection{\textsc{*Shuffle} Languages.}

The first set of impossible languages, which we call the \textsc{*Shuffle} languages, involve different shuffles of tokenized English sentences. 

\begin{enumerate}
    \item \textbf{\textcolor{controlshuf}{\textsc{NoShuffle}}}: The input sentence is tokenized, and the token sequence is unaltered. This language is simply English, used for comparison with other \textsc{*Shuffle} languages.
    
    \item \textbf{\textcolor{nondetshuf}{\textsc{NondeterministicShuffle}}}: The tokenized input sentence is randomly shuffled. A different random shuffle is used for each input sentence, with no consistency across inputs.

    \item \textbf{\textcolor{detshuf57}{\textsc{DeterministicShuffle}($s$)}}: The tokenized input sentence is deterministically shuffled based on the length of the token sequence. For example, all token sequences of length~5 are shuffled in the same order. We create several languages by varying the random seed~$s$ that produces the shuffle.

    \item \textbf{\textcolor{locshuf5}{\textsc{LocalShuffle}($w$)}}: The tokenized input sentence is deterministically shuffled in local windows of a fixed size $w$. We create several languages by varying $w$.

    \item \textbf{\textcolor{evenodd}{\textsc{EvenOddShuffle}}}: The tokenized input sentence is reordered such that all even-indexed tokens appear first, followed by all odd-indexed tokens.
    
\end{enumerate}

\vspace{6pt}
\noindent
The random shuffling function that generates the \textsc{NondeterministicShuffle} language is irreversible, resulting in sentences that are purely bags of words---any structural information in the original linguistic signal is irretrievable. 
While the \textsc{DeterministicShuffle} languages are created using a reversible perturbation function, this function operates in an entirely non-linguistic manner; words are ordered based solely on the random seed and sentence length, without considerations for linguistic features or \emph{information locality}---the property that, when parts of text predict each other, they are often close together \citep{futrell-2019-information,mansfield2023emergence}. This method is arguably even less humanly feasible than \textsc{NondeterministicShuffle}, as it relies on an arbitrarily complex yet consistent rule to determine word order.\footnote{Even in the imaginable case of a language with completely free word order, it seems extremely unlikely that this freedom would be totally insensitive to any clause boundaries while the language otherwise looks morphologically like English does. It thus seems very safe to assume that our \textsc{NondeterministicShuffle} language counts as impossible.} 
The question of ranking these two families of languages in the impossibility continuum probes at the definition of impossibility and whether reversibility to an attested language like English is a relevant quantity.

The \textsc{LocalShuffle} languages offer a finer-grained testbed for the importance of information locality, since we can observe the effects of different window sizes. Finally, \textsc{EvenOddShuffle} also manipulates locality, but interestingly preserves part of the linear word order of English while introducing new long-distance dependencies.

\subsection{\textsc{*Reverse} Languages.}

The \textsc{*Reverse} impossible languages involve reversals of all or part of input sentences.

\begin{enumerate}
    \item \textbf{\textcolor{controlrev}{\textsc{NoReverse}}}: The input sentence is tokenized, and a special marker token \revmarker\space is inserted at a random position in the token list. Like \textsc{NoShuffle}, this language is most similar to English. We use it for comparison with other \textsc{*Reverse} languages.

    \item \textbf{\textcolor{partialrev}{\textsc{PartialReverse}}}: The input sentence is  tokenized, a special marker token \revmarker\space is inserted at a random position in the list of tokens, and the following tokens are reversed.

    \item \textbf{\textcolor{fullrev}{\textsc{FullReverse}}}: The input sentence is tokenized, a special marker token \revmarker\space is inserted at a random position in the token list, and \emph{all} tokens are reversed.
    
\end{enumerate}

\vspace{6pt}
\noindent
The \textsc{PartialReverse} language is inspired by the experiments of \citet{mitchell-bowers-2020-priorless} on partially reversed English data, though our experiments are not a direct replication, since we use a different model architecture and dataset. \textsc{FullReverse} may seem like a plausible language syntactically, but higher-level linguistic concepts like anaphora would be highly disrupted.
The \revmarker\space tokens are placed at the same positions across the data in all \textsc{*Reverse} languages to control for the entropy introduced by their random placement.

\subsection{\textsc{*Hop} Languages.}

The \textsc{*Hop} languages perturb verb inflection with counting rules.

\begin{enumerate}
    \item \textbf{\textcolor{NoHop}{\textsc{NoHop}}}: All 3rd-person present tense verbs in the input sentence are lemmatized, and the sentence is tokenized. For each 3rd-person present tense verb, a special marker representing the verb's number and tense is placed right after the lemmatized verb. Singular verbs are marked with a special token \singularmarker, and plural verbs are marked with \pluralmarker. Like the other control languages, \textsc{NoHop} has a pattern that is most similar to English.

    \item \textbf{\textcolor{tokenhop}{\textsc{TokenHop}}}: Identical transformation to \textsc{NoHop}, but the special number/tense markers are placed 4 tokens after the verb.

    \item \textbf{\textcolor{wordhop}{\textsc{WordHop}}}: Identical transformation to \textsc{NoHop} and \textsc{TokenHop}, but the special number/tense markers are placed 4 \emph{words} after the verb, skipping punctuation.
    
\end{enumerate}

\vspace{6pt}
\noindent
These languages specifically investigate GPT-2's ability to learn grammar rules that involve counting the positions of words or tokens.

\section{Experiments}

\begin{figure*}
    \centering
     \includegraphics[width=1.0\textwidth]{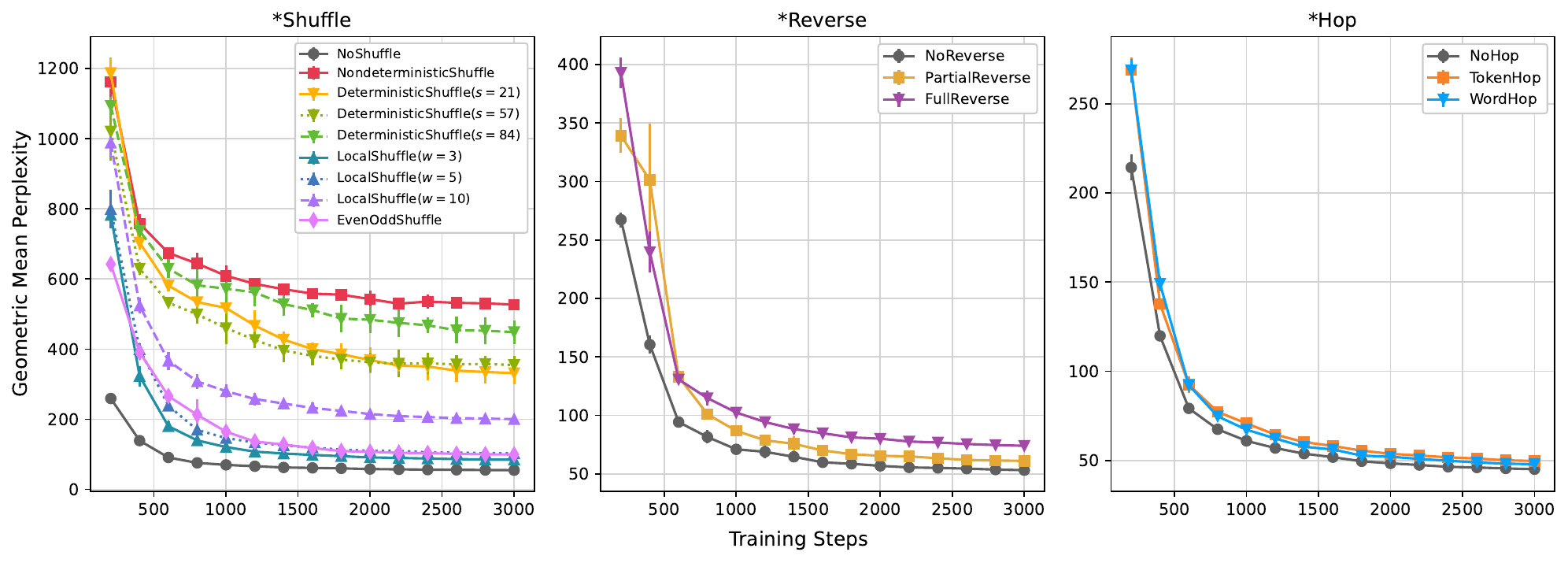}
    \caption{Perplexities on a sample of 10K test sentences for each
    impossible language model over training steps. Error bars indicate 95\% confidence intervals across 5 training runs initialized with different random seeds and evaluated on different test samples.}
    \label{fig:perplexities}
\end{figure*}

We run several experiments to assess GPT-2's learning of our impossible languages. Our first experiment (\Cref{sec:perplexities}) uses perplexities as a general evaluation to compare how well each impossible language model has learned its own perturbed language and see whether this reflects the hypothesized impossibility continuum. In our second and third experiments, we conduct a closer examination of the \textsc{*Hop} languages. Given that their count-based verb marking rules appear to be the least clearly implausible among our proposed languages, we focus on examining these rules specifically through targeted assessments using surprisal theory (\Cref{sec:surprisals}). Finally, we dive deeper into the mechanisms each \textsc{*Hop} model uses to predict their respective verb marking rules using causal abstraction analysis (\Cref{sec:causal-abstractions}). For all evaluations, we run tests on several model checkpoints to observe the learning process over intervals of training steps.\footnote{
We also conduct a constituency probing experiment to test effects on GPT-2's implicit understanding of syntax, with minimal observed differences among models (see \Cref{sec:appendix-probing}).}

\subsection{Implementation Details}

For each impossible language, we apply its perturbation function to each sentence of the BabyLM dataset \cite{warstadt2023papers} to create a transformed dataset. 
\Cref{sec:appendix-dataset-filters} provides details on preprocessing and formatting, and describes the language-specific filtering needed to achieve the criteria that define each language.

We train standard GPT-2 small models \cite{radford2018improving, radford2019languagemodels} on each impossible language. To produce confidence intervals for our experiments, we train 5 sets of models for each language using different random seeds, which affect the model parameter initialization and dataset shuffling during training. Training and model hyperparameter choices are detailed in \Cref{sec:appendix-hyperparameters}.
The primary set of GPT-2 models we train have absolute positional encodings. We also train a set of GPT-2 small models with an architecture in which the positional encodings are removed, so that the models' only notion of word order is derived from \mbox{GPT-2's} causal language modeling learning objective \cite{kazemnejad2023impact}. Results for these additional experiments supported our main findings on the unaltered GPT-2 architecture. These results are provided in \Cref{sec:appendix-no-pos-encoding}.

\subsection{Experiment 1: Language Models Reflect the Impossibility Continuum}
\label{sec:perplexities}

We train GPT-2 models on all of the languages described in \Cref{tab:perturbations}, and evaluate each model's perplexities on a test set over the course of training. Test perplexities provide a general metric for the extent to which a model has learned a language.

\paragraph{Setup.}
We sample 10K sentences from the BabyLM test set and perturb this sample for each impossible language. For a given impossible language model, we report the geometric mean of the individual sentence perplexities in the corresponding test sample.

\paragraph{Hypothesis.}  Models trained on possible languages will achieve lower average perplexities more quickly (as measured in training steps) than those trained on impossible languages.

\paragraph{Results.} Our results are in \Cref{fig:perplexities}.
There are clear distinctions between model perplexities after about 500 training steps. First considering the \textsc{*Shuffle} models, the \textsc{NondeterministicShuffle} model has the highest perplexities, followed by the three \textsc{DeterministicShuffle} models, indicating that GPT-2 is better at learning shuffling patterns when they are deterministic, invertible functions.\footnote{This result is also supported by separate evaluations of each \textsc{DeterministicShuffle} model on test data from other shuffles (see \Cref{sec:appendix-deterministic-shuffle}). Each model has lower perplexities on its own deterministic shuffle.} 
The prevalence of certain sentence lengths in the corpus could also limit the variety of sentence shuffles in the \textsc{DeterministicShuffle} languages, potentially resulting in similarly functioning words frequently occupying the same token positions, thus increasing their predictability.

Following the sentence-level shuffles, the next models in the order of decreasing perplexity are the three \textsc{LocalShuffle} models, with smaller window sizes having lower perplexities. \textsc{LocalShuffle}$(w=3)$ and \textsc{EvenOddShuffle} have perplexities closest to the \textsc{NoShuffle} model (which represents unaltered English), but \textsc{NoShuffle} consistently has the lowest perplexities throughout the training process.

Compared to the \textsc{*Shuffle} models, the experimental \textsc{*Reverse} models have perplexities that are much closer to the \textsc{NoReverse} model, and \textsc{PartialReverse} is slightly better than \textsc{FullReverse}. For the \textsc{*Hop} languages, their respective control model again has the lowest perplexities, although differences among the models are quite minimal.
This warrants our deep-dive into the particular verb marking patterns for this set of models.

\subsection{Experiment 2: Language Models Disprefer Counting Rules} \label{sec:surprisals}

\begin{figure*}
    \centering
    \begin{subfigure}[t]{0.48\textwidth}
        \centering
        \includegraphics[width=1.0\linewidth]{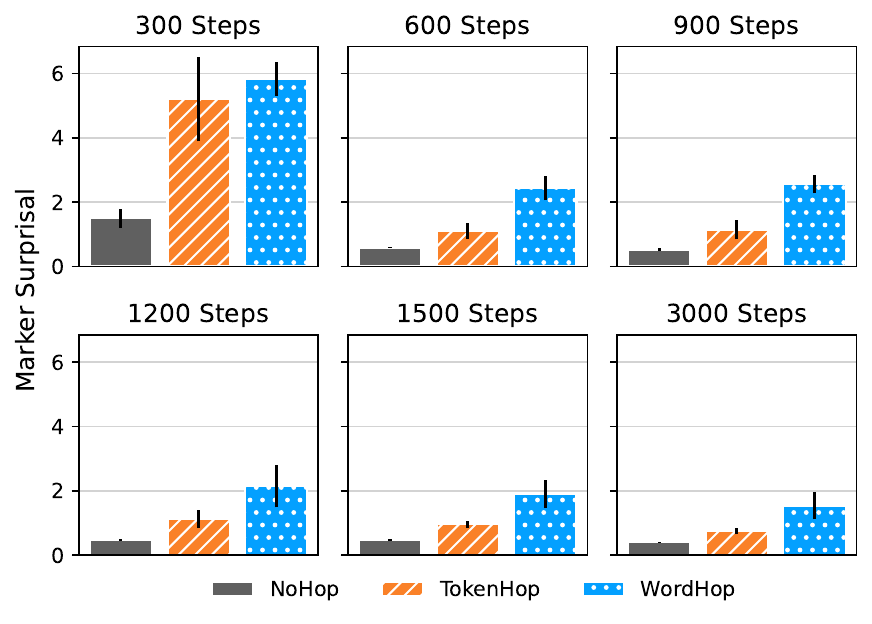}
        \caption{Test 1: mean surprisals of the verb marker token (\singularmarker
        \space or \pluralmarker) for each \textsc{*Hop} model. }
        \label{fig:surprisals}
    \end{subfigure}\hfill
    \begin{subfigure}[t]{0.48\textwidth}
        \centering
        \includegraphics[width=1.0\linewidth]{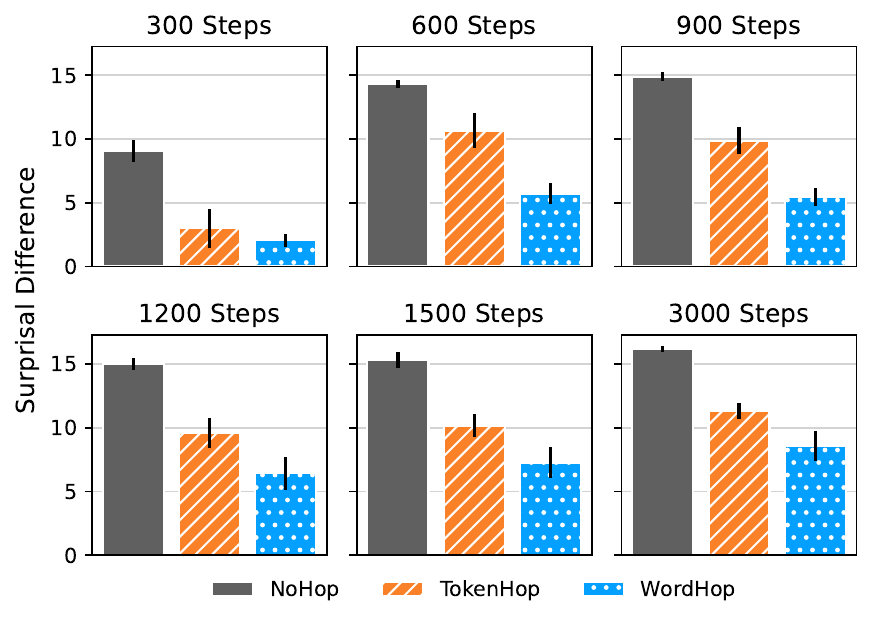}
        \caption{Test 2: mean surprisal difference between the verb marker token (\singularmarker
        \space or \pluralmarker) and the following token for each \textsc{*Hop} model.}
        \label{fig:surprisal_differences}
    \end{subfigure}
    \caption{Surprisal tests for each \textsc{*Hop} model over training steps. Error bars indicate 95\% confidence intervals across 5 training runs initialized with different random seeds and evaluated on different test samples.}
    \label{fig:surprisal_results}
\end{figure*}

In Experiment 1, we show that impossible languages are harder for GPT-2 to learn. However, perplexity is a coarse-grained metric of language learning, and the question remains: do language models learn natural grammatical structures better than impossible grammars?

The structure of the \textsc{*Hop} languages invites a
finer-grained evaluation of their verb marking rules. We use \emph{surprisals} to measure how well each \textsc{*Hop} model can predict the placement of its
verb marker tokens, \singularmarker \space and \pluralmarker. The surprisal
$S(w_i)$ of a word $w_i$ is the negative log probability of $w_i$ given
the context words $w_1, \ldots, w_{i-1}$ that precede it:
$S(w_i) = -\log_2 p(w_i|w_1, \ldots,w_{i-1})$.
Surprisals have been used as acceptability judgments from neural language models to probe for their processing of syntactic information 
\cite{wilcox-etal-2018-rnn, futrell-etal-2019-neural, hu-etal-2020-systematic, 
wilcox2023syntactic} and have been shown to correlate with human sentence processing difficulty
\cite{hale-2001-probabilistic, levy2008expectation}.

\paragraph{Setup.}
To test the \textsc{*Hop} models' sensitivity to marker placement, we conduct two tests on a sample of 10K sentences extracted from the BabyLM dataset containing the verb marker tokens (\singularmarker \space or \pluralmarker).
As an example, consider the following pair of sentences for the \textsc{NoHop} language shown in~\ref{ex:control_hop_pair}.

\ex. \label{ex:control_hop_pair} 
\a. \textsmaller[1]{\texttt{\tokenHe\,\tokenclean\,\tokensing\,\tokenhis\,\tokenvery\,\tokenmessy\,\tokenbooks\,\tokenhe\,\tokenlf\,\tokenperiod}} \label{ex:control_hop_a}
\b. *\textsmaller[1]{\texttt{\tokenHe\,\tokenclean}\_\_\texttt{\tokenhis\,\tokenvery\,\tokenmessy\,\tokenbooks\,\tokenhe\,\tokenlf\,\tokenperiod}} \label{ex:control_hop_b}

Sentence~\ref{ex:control_hop_a} is an example in the \textsc{NoHop} language, and \ref{ex:control_hop_b} is an ungrammatical counterfactual in which the marker token does not appear.

In the first test, we compare the average surprisals of the marker tokens across the three \textsc{*Hop} languages, using grammatical examples like~\ref{ex:control_hop_a}.
In the case of~\ref{ex:control_hop_a}, the marker is singular, and its surprisal $S(\texttt{\,\tokensing\,})$ is defined as: 
$$S(\texttt{\,\tokensing\,}) = -\log_2p(\texttt{\,\tokensing\,$|$\,\tokenHe\,\tokenclean\,})$$
We average this surprisal value for instances of \singularmarker \space or \pluralmarker \space in the test sample.

In the second test, we construct minimal pairs from the example sentences in which the marker token appears and does not appear, and then compare the surprisal of the marker token to the surprisal of the token that follows it, both conditioned on the same context. 
In example~\ref{ex:control_hop_b}, the surprisal of the following token $S(\texttt{\,\tokenhis\,})$ is defined as:
$$S(\texttt{\,\tokenhis\,}) = -\log_2p(\texttt{\,\tokenhis\,$|$\,\tokenHe\,\tokenclean\,})$$
We expect $S(\texttt{\,\tokenhis\,}) - S(\texttt{\,\tokensing\,})$
to be a large positive value.
We average such surprisal differences over instances of the marker tokens in the test sample and similarly define marker surprisals and minimal pair configurations for the other \textsc{*Hop} languages.

\paragraph{Hypothesis.} 
For the first surprisal test, our hypothesis is that the mean surprisal of the marker tokens across test examples will be smaller for the control language than for the impossible languages.
For the second test, our hypothesis is that the mean surprisal difference across all test pairs will be larger for possible languages than for impossible ones.

\paragraph{Results.}

Our results are presented in \Cref{fig:surprisal_results}. The \textsc{NoHop} model, which has the verb marking pattern most similar to English, consistently has the lowest mean marker surprisal across training steps in test 1 (\Cref{fig:surprisals}). The \textsc{NoHop} model also has the highest mean surprisal difference across training steps in test 2 (\Cref{fig:surprisal_differences}). Both of these results indicate that GPT-2 has learned to expect the marker tokens when they follow a more natural grammatical pattern and was very surprised when they did not appear at the correct positions.

GPT-2 learns to expect marker tokens at the right locations in the other \textsc{*Hop} models, just not as well as the control. \textsc{TokenHop} tends to have a lower marker surprisal and a higher mean surprisal difference compared to \textsc{WordHop} across training steps, indicating that GPT-2 is better at learning the verb marking rule when the units being counted are tokens instead of words.

\subsection{Experiment 3: Language Models Develop Natural Solutions to Unnatural Patterns} \label{sec:causal-abstractions}

Experiment 2 demonstrates that, while GPT-2 favors natural grammar rules, it is also capable of acquiring count-based grammar rules like those seen in the verb marking patterns of our \textsc{*Hop} languages. But what sorts of internal mechanisms does it implement to learn such grammar rules, and how do these mechanisms compare to the more natural control? To address this, we conduct a final experiment using \emph{causal abstraction analysis}, which offers an interpretability framework for identifying and examining causal mechanisms within neural models \cite{geiger-etal-2020-neural, geiger2021causal, wu-etal-2022-causal, pmlr-v202-wu23b, wu-etal-2023-Boundless-DAS, geiger2023finding}. We employ the \emph{interchange intervention} technique on our \textsc{*Hop} models. To perform a basic interchange intervention on a neural model $M$, we create two instances of $M$ that are provided two different inputs, the base input $b$ and the source input~$s$. Then, we interchange representations created while processing~$b$ with representations created while processing~$s$ and observe the effect on the output of~$M$. Such interventions allow us to piece together a causal understanding of how the model processes inputs.

\paragraph{Setup.}

\begin{figure}
\centering
\includegraphics[width=0.4\textwidth]{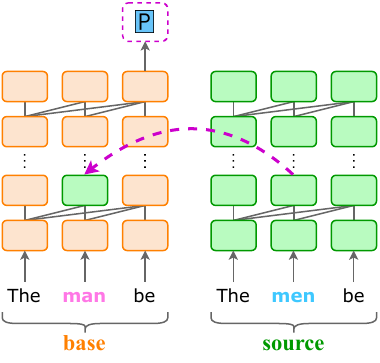}
\caption{An \emph{interchange intervention} on the \textsc{NoHop} model with base input $b = \texttt{The man be}$ and source input $s = \texttt{The men be}$. The intervention is performed at the second layer and second token position, causing a change in prediction from \tokensing \space to \tokenplur.}
\label{fig:hop-intervention}
\end{figure}

\begin{figure*}[ht]
     \centering
     \includegraphics[width=0.329\textwidth]{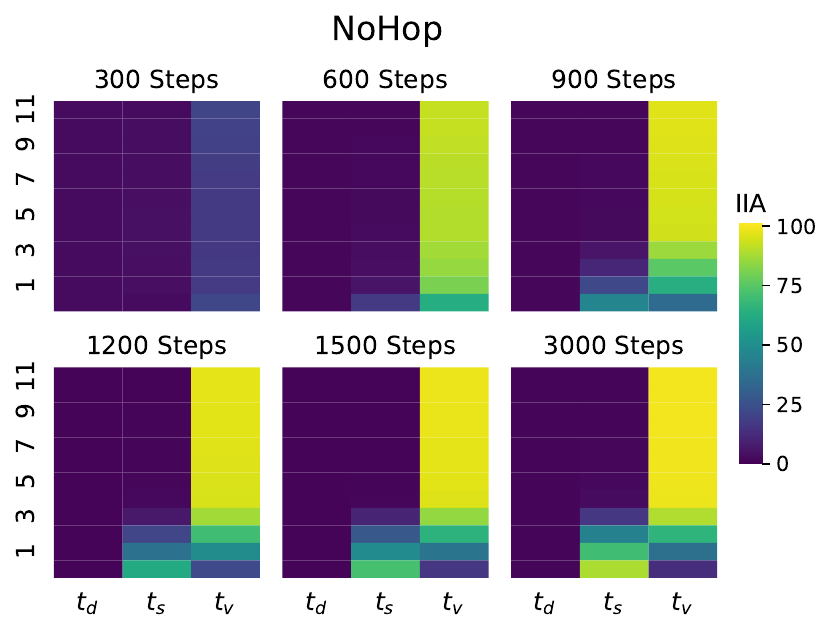}
     \includegraphics[width=0.329\textwidth]{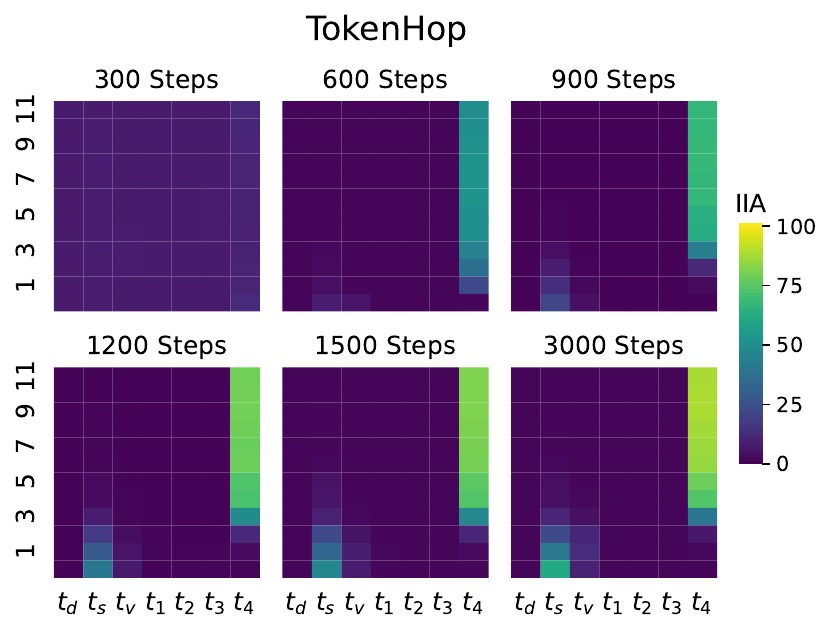}
     \includegraphics[width=0.329\textwidth]{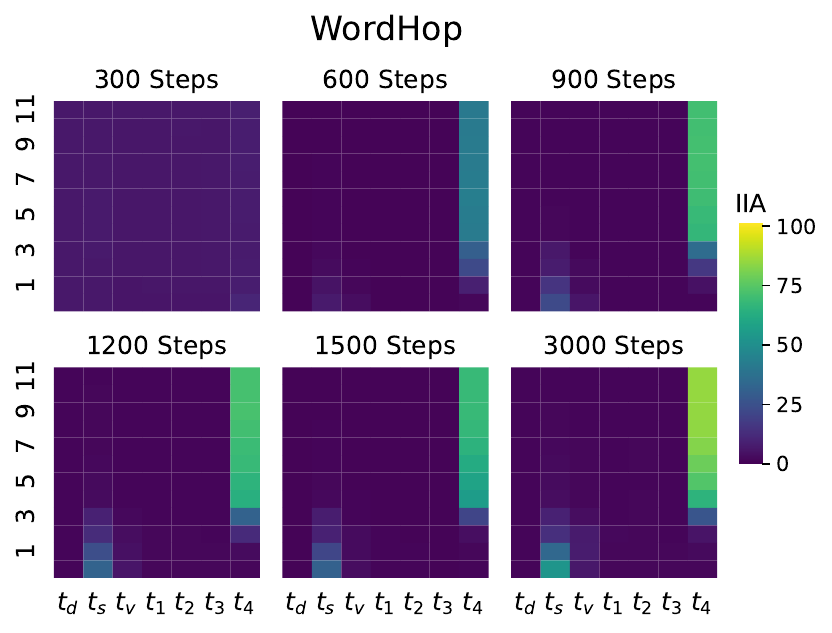}
    \caption{Subject--verb agreement interchange intervention accuracies (IIA) for each
    \textsc{*Hop} model over training steps. Vertical axes denote the GPT-2 layer of the intervention, and horizontal axes denote the token position of the intervention. $t_d$, $t_s$, and $t_v$ represent the tokens for the determiner, subject, and verb, respectively. $t_1 \dots t_4$ represent the four tokens/words between the verb and its marker for \textsc{TokenHop} and \textsc{WordHop}.
    IIA values are averaged over results from 5 models initialized on different random seeds. See \Cref{sec:intervention_confidence_intervals} for confidence intervals.
    }
    \label{fig:iia_hop_models}
\end{figure*}

We use interchange interventions to identify representations in our \textsc{*Hop} models 
that have causal effects on their output behaviors on a subject--verb agreement task. In 
our experimental setup, $b$ is a sentence prefix with a singular subject and $s$ is
an identical prefix with the plural form of the subject. These prefixes include all
tokens up to but not including the markers (\singularmarker \space and \pluralmarker).
We interchange the GPT-2 block outputs from processing~$b$ with GPT-2 block outputs
from processing~$s$ and observe %
whether the probability of plural marker~\pluralmarker \space is higher than the probability
of singular marker~\singularmarker \space after the intervention. This is 
shown more concretely in \Cref{fig:hop-intervention}. 

We run such interventions at each GPT-2 layer and token position to see which 
parts of the model cause a change in the marker prediction. We run all of these
interventions over several test examples and report the \emph{interchange intervention
accuracy} (IIA), a metric that represents the subject--verb agreement accuracy if the
counterfactual (i.e.\ plural) were the ground truth. The test examples for each \textsc{*Hop} model are extracted from their respective versions of the BabyLM test set, and minimally-different counterfactual examples are created by changing the singular subjects to plural subjects.
To ensure that interventions on different examples are analogous, we use regular expressions to locate examples that follow the same structure (i.e.\ subjects and verbs at the same positions).

\paragraph{Results.}

Our results are presented in \Cref{fig:iia_hop_models}. The IIA graphs
demonstrate how information about the marker tokens flows through the models.
We can see that, in all three \textsc{*Hop} models, IIA is high at the token position 
of the subject up until about layer 3; then there is a transition to the position of
the last token in the prefix, preceding the location where the marker should be predicted.
All models develop the same modular solution to the task by tracking
agreement through the representations at the relevant positions, but the
\textsc{NoHop} model obtains nearly 100\% IIA earlier during training, at about
1,500 training steps, supporting the previous surprisal results.

\section{Discussion and Conclusion}

Contra claims by Chomsky and others that LLMs cannot possibly inform our understanding of human language, we argue there is great value in treating LLMs as a comparative system for human language and in understanding what systems like LLMs can and cannot learn. 
Prior explorations of neural language models have already been fruitful for understanding the generalization of syntactic principles from data \cite{wilcox-etal-2018-rnn, marvin-linzen-2018-targeted, futrell-etal-2019-neural, prasad-etal-2019-using, hu-etal-2020-systematic}. Our paper complements this line of work.
We have shown that GPT-2 models do not master our set of synthetic impossible languages as well as natural ones, challenging the unfounded assertions stated previously.

Even in the absence of a clear definition of what constitutes a possible or impossible language, we believe that our investigations advance this debate regarding LLMs. The lack of a definition does not hinder inquiry into this topic; in fact, it beckons further explorations of the boundary between the possible and impossible languages, as shown in our hypothesized continuum in \Cref{fig:language-continuum}. We believe that the \textsc{*Hop} languages we propose closely approach this boundary.

At the same time, conclusions about LLMs' linguistic competence and preferences for natural languages should be informed by an understanding of the ways that models fundamentally differ from humans. For instance, we saw that models can perform operations that involve counting tokens because LLMs rely on tokens as basic units. While humans are sensitive to morpheme boundaries and word boundaries, it is unlikely humans rely on atomic tokens in the way that LLMs do.
This does not mean that LLMs can fundamentally tell us nothing about human language.
Rather, as we did here, it is valuable to consider and control for this difference before making generalizations.

Since at least the 1950s, a major line of linguistic inquiry has focused on what aspects of syntactic structure can be learned just from data, without domain-specific innate priors (e.g.~a \emph{Universal Grammar}).
LLMs lack strong in-built linguistic priors, yet they can learn complex syntactic structures.
While many LLMs are trained with vastly more data than children see, there is increasing evidence that even systems trained on smaller amounts of data can learn interesting linguistic information \citep{warstadt2023papers}.
The current paper raises further questions along similar lines. Since we do find that real languages are more learnable by GPT-2, this leads us to wonder what inductive bias of GPT language models matches natural language.
We believe that this inductive bias is related to information locality, the tendency for statistical correlations in text to be short range.
Information locality arises in GPTs due to their autoregressive training objective and has been argued to arise in humans due to the incremental nature of real-time language processing \citep{futrell-2019-information,Hahn2021}.

Since LLMs have been shown to learn the complex structures of human language and have a preference for learning such structures over unnatural counterfactuals, it follows that they are clearly relevant to investigations and claims about the necessary innate priors for language learning.
Arguments that they are ``by design, unlimited in what they can `learn'''  and ``incapable of distinguishing the possible from the impossible'' \citep{chomsky2023nyt} do not offer convincing evidence otherwise.

\section{Acknowledgments}

The authors would like to thank Aryaman Arora, Christiane Fellbaum, Roger Levy, Tristan Thrush, and Diyi Yang for helpful comments on the project. We would also like to thank the members of the Stanford NLP Group, the MIT Computational Psycholinguistics Lab, and the anonymous reviewers for useful discussions. Julie Kallini is supported by a National Science Foundation Graduate Research Fellowship under grant number DGE-2146755.

\section{Limitations}

Due to resource constraints, we exclusively use the GPT-2 architecture to train models on our various synthetic impossible languages. Each of our experiments involves training a GPT-2 model from scratch on a different language dataset, and for every such language, we train multiple GPT-2 models to establish confidence intervals for our evaluation metrics. Applying this approach to several different model architectures would be quite resource-intensive, so we opted to choose a single architecture in this paper. Future work could apply our methodology to models trained with different architectures or training objectives.

Our impossible languages are derived by manipulating an English dataset. While we do not conduct experiments that use other natural languages as a starting point, our experimental choices (i.e.\ the synthetic languages we design) are informed by linguistic diversity and typology, distinguishing our impossible languages from those that are rare but attested. However, future work might involve deriving impossible languages from base languages other than English and include more morphological manipulations.

\section{Ethics Statement}

While this work makes the case for language models as useful tools for cognitive science and linguistics research, these models learn and generate language through processes that are fundamentally different from those employed by humans. Making direct claims about human language learning based on the results of this paper could pose potential risks and harms. This research merely aims to explore the learnability of different languages (specifically, those languages that \emph{cannot} be acquired by humans and are not representative of any known human language) through the lens of neural models. 

\bibliography{anthology,custom}

\begin{thebibliography}{69}
\expandafter\ifx\csname natexlab\endcsname\relax\def\natexlab#1{#1}\fi

\bibitem[{Abdou et~al.(2022)Abdou, Ravishankar, Kulmizev, and S{\o}gaard}]{abdou-etal-2022-word}
Mostafa Abdou, Vinit Ravishankar, Artur Kulmizev, and Anders S{\o}gaard. 2022.
\newblock \href {https://doi.org/10.18653/v1/2022.acl-long.476} {Word order does matter and shuffled language models know it}.
\newblock In \emph{Proceedings of the 60th Annual Meeting of the Association for Computational Linguistics (Volume 1: Long Papers)}, pages 6907--6919, Dublin, Ireland. Association for Computational Linguistics.

\bibitem[{Alleman et~al.(2021)Alleman, Mamou, A~Del~Rio, Tang, Kim, and Chung}]{alleman-etal-2021-syntactic}
Matteo Alleman, Jonathan Mamou, Miguel A~Del~Rio, Hanlin Tang, Yoon Kim, and SueYeon Chung. 2021.
\newblock \href {https://doi.org/10.18653/v1/2021.repl4nlp-1.27} {Syntactic perturbations reveal representational correlates of hierarchical phrase structure in pretrained language models}.
\newblock In \emph{Proceedings of the 6th Workshop on Representation Learning for NLP (RepL4NLP-2021)}, pages 263--276, Online. Association for Computational Linguistics.

\bibitem[{Bolhuis et~al.(2024)Bolhuis, Crain, Fong, and Moro}]{Bolhuis2024}
Johan~J. Bolhuis, Stephen Crain, Sandiway Fong, and Andrea Moro. 2024.
\newblock \href {https://doi.org/10.1038/d41586-024-00824-z} {Three reasons why {AI} doesn't model human language}.
\newblock \emph{Nature}, 627(8004):489–489.

\bibitem[{Chomsky(1956)}]{chomsky1956models}
Noam Chomsky. 1956.
\newblock \href {https://doi.org/10.1109/TIT.1956.1056813} {Three models for the description of language}.
\newblock \emph{IRE Transactions on Information Theory}, 2(3):113--124.

\bibitem[{Chomsky(1957)}]{chomsky1957syntactic}
Noam Chomsky. 1957.
\newblock \href {https://doi.org/doi:10.1515/9783112316009} {\emph{Syntactic Structures}}.
\newblock De Gruyter Mouton, Berlin, Boston.

\bibitem[{Chomsky(1959)}]{chomsky1959certain}
Noam Chomsky. 1959.
\newblock \href {https://doi.org/https://doi.org/10.1016/S0019-9958(59)90362-6} {On certain formal properties of grammars}.
\newblock \emph{Information and Control}, 2(2):137--167.

\bibitem[{Chomsky(1965)}]{chomsky1965aspects}
Noam Chomsky. 1965.
\newblock \emph{Aspects of the Theory of Syntax}.
\newblock The MIT Press.

\bibitem[{Chomsky(2002)}]{chomsky2002nature}
Noam Chomsky. 2002.
\newblock \href {https://doi.org/10.1017/CBO9780511613876} {\emph{On Nature and Language}}.
\newblock Cambridge University Press.

\bibitem[{Chomsky(2023)}]{chomsky2023cowen}
Noam Chomsky. 2023.
\newblock \href {https://conversationswithtyler.com/episodes/noam-chomsky/} {Conversations with {Tyler}: {Noam} {Chomsky}}.
\newblock Conversations with {Tyler} Podcast.

\bibitem[{Chomsky et~al.(2023)Chomsky, Roberts, and Watumull}]{chomsky2023nyt}
Noam Chomsky, Ian Roberts, and Jeffrey Watumull. 2023.
\newblock \href {https://www.nytimes.com/2023/03/08/opinion/noam-chomsky-chatgpt-ai.html} {Noam {Chomsky}: The false promise of {ChatGPT}}.
\newblock \emph{The New York Times}.

\bibitem[{Comrie(1989)}]{comrie1989language}
Bernard Comrie. 1989.
\newblock \emph{Language universals and linguistic typology: Syntax and morphology}.
\newblock University of Chicago press.

\bibitem[{Deletang et~al.(2023)Deletang, Ruoss, Grau-Moya, Genewein, Wenliang, Catt, Cundy, Hutter, Legg, Veness, and Ortega}]{deletang2023neural}
Gregoire Deletang, Anian Ruoss, Jordi Grau-Moya, Tim Genewein, Li~Kevin Wenliang, Elliot Catt, Chris Cundy, Marcus Hutter, Shane Legg, Joel Veness, and Pedro~A Ortega. 2023.
\newblock \href {https://openreview.net/forum?id=WbxHAzkeQcn} {Neural networks and the {Chomsky} hierarchy}.
\newblock In \emph{The Eleventh International Conference on Learning Representations}.

\bibitem[{Devlin et~al.(2019)Devlin, Chang, Lee, and Toutanova}]{devlin-etal-2019-bert}
Jacob Devlin, Ming-Wei Chang, Kenton Lee, and Kristina Toutanova. 2019.
\newblock \href {https://doi.org/10.18653/v1/N19-1423} {{BERT}: Pre-training of deep bidirectional transformers for language understanding}.
\newblock In \emph{Proceedings of the 2019 Conference of the North {A}merican Chapter of the Association for Computational Linguistics: Human Language Technologies, Volume 1 (Long and Short Papers)}, pages 4171--4186, Minneapolis, Minnesota. Association for Computational Linguistics.

\bibitem[{Ebrahimi et~al.(2020)Ebrahimi, Gelda, and Zhang}]{ebrahimi-etal-2020-self}
Javid Ebrahimi, Dhruv Gelda, and Wei Zhang. 2020.
\newblock \href {https://doi.org/10.18653/v1/2020.findings-emnlp.384} {How can self-attention networks recognize {D}yck-n languages?}
\newblock In \emph{Findings of the Association for Computational Linguistics: EMNLP 2020}, pages 4301--4306, Online. Association for Computational Linguistics.

\bibitem[{Elman(1990)}]{elman1990finding}
Jeffrey~L. Elman. 1990.
\newblock \href {https://doi.org/https://doi.org/10.1016/0364-0213(90)90002-E} {Finding structure in time}.
\newblock \emph{Cognitive Science}, 14(2):179--211.

\bibitem[{Evans and Levinson(2009)}]{evans2009myth}
Nicholas Evans and Stephen~C Levinson. 2009.
\newblock The myth of language universals: Language diversity and its importance for cognitive science.
\newblock \emph{Behavioral and brain sciences}, 32(5):429--448.

\bibitem[{Everett(2012)}]{everett2012piraha}
Daniel~L. Everett. 2012.
\newblock \href {https://doi.org/https://doi.org/10.1002/wcs.1195} {What does {Pirahã} grammar have to teach us about human language and the mind?}
\newblock \emph{WIREs Cognitive Science}, 3(6):555--563.

\bibitem[{Futrell(2019)}]{futrell-2019-information}
Richard Futrell. 2019.
\newblock \href {https://doi.org/10.18653/v1/W19-7902} {Information-theoretic locality properties of natural language}.
\newblock In \emph{Proceedings of the First Workshop on Quantitative Syntax (Quasy, SyntaxFest 2019)}, pages 2--15, Paris, France. Association for Computational Linguistics.

\bibitem[{Futrell and Hahn(2022)}]{futrell2022information}
Richard Futrell and Michael Hahn. 2022.
\newblock \href {https://doi.org/10.3389/fcomm.2022.657725} {Information theory as a bridge between language function and language form}.
\newblock \emph{Frontiers in Communication}, 7.

\bibitem[{Futrell et~al.(2019)Futrell, Wilcox, Morita, Qian, Ballesteros, and Levy}]{futrell-etal-2019-neural}
Richard Futrell, Ethan Wilcox, Takashi Morita, Peng Qian, Miguel Ballesteros, and Roger Levy. 2019.
\newblock \href {https://doi.org/10.18653/v1/N19-1004} {Neural language models as psycholinguistic subjects: Representations of syntactic state}.
\newblock In \emph{Proceedings of the 2019 Conference of the North {A}merican Chapter of the Association for Computational Linguistics: Human Language Technologies, Volume 1 (Long and Short Papers)}, pages 32--42, Minneapolis, Minnesota. Association for Computational Linguistics.

\bibitem[{Galke et~al.(2023)Galke, Ram, and Raviv}]{galke2023makes}
Lukas Galke, Yoav Ram, and Limor Raviv. 2023.
\newblock \href {http://arxiv.org/abs/2302.12239} {What makes a language easy to deep-learn?}

\bibitem[{Geiger et~al.(2021)Geiger, Lu, Icard, and Potts}]{geiger2021causal}
Atticus Geiger, Hanson Lu, Thomas Icard, and Christopher Potts. 2021.
\newblock \href {https://proceedings.neurips.cc/paper_files/paper/2021/file/4f5c422f4d49a5a807eda27434231040-Paper.pdf} {Causal abstractions of neural networks}.
\newblock In \emph{Advances in Neural Information Processing Systems}, volume~34, pages 9574--9586. Curran Associates, Inc.

\bibitem[{Geiger et~al.(2020)Geiger, Richardson, and Potts}]{geiger-etal-2020-neural}
Atticus Geiger, Kyle Richardson, and Christopher Potts. 2020.
\newblock \href {https://doi.org/10.18653/v1/2020.blackboxnlp-1.16} {Neural natural language inference models partially embed theories of lexical entailment and negation}.
\newblock In \emph{Proceedings of the Third BlackboxNLP Workshop on Analyzing and Interpreting Neural Networks for NLP}, pages 163--173, Online. Association for Computational Linguistics.

\bibitem[{Geiger et~al.(2023)Geiger, Wu, Potts, Icard, and Goodman}]{geiger2023finding}
Atticus Geiger, Zhengxuan Wu, Christopher Potts, Thomas Icard, and Noah~D. Goodman. 2023.
\newblock Finding alignments between interpretable causal variables and distributed neural representations.
\newblock In \emph{Proceedings of Causal Learning and Reasoning 2024.}

\bibitem[{Greenberg(1963)}]{greenberg1963universals}
Joseph Greenberg. 1963.
\newblock \href {https://doi.org/https://doi.org/10.1002/wcs.1195} {Some universals of grammar with particular reference to the order of meaningful elements}.
\newblock \emph{Universals of Language}, pages 73--113.

\bibitem[{Hahn(2020)}]{hahn2020theoretical}
Michael Hahn. 2020.
\newblock \href {https://doi.org/10.1162/tacl_a_00306} {Theoretical limitations of self-attention in neural sequence models}.
\newblock \emph{Transactions of the Association for Computational Linguistics}, 8:156--171.

\bibitem[{Hahn et~al.(2021)Hahn, Degen, and Futrell}]{Hahn2021}
Michael Hahn, Judith Degen, and Richard Futrell. 2021.
\newblock \href {https://doi.org/10.1037/rev0000269} {Modeling word and morpheme order in natural language as an efficient trade-off of memory and surprisal.}
\newblock \emph{Psychological Review}, 128(4):726–756.

\bibitem[{Hahn et~al.(2020)Hahn, Jurafsky, and Futrell}]{hahnetal2020universals}
Michael Hahn, Dan Jurafsky, and Richard Futrell. 2020.
\newblock \href {https://doi.org/10.1073/pnas.1910923117} {Universals of word order reflect optimization of grammars for efficient communication}.
\newblock \emph{Proceedings of the National Academy of Sciences}, 117(5):2347--2353.

\bibitem[{Hale(2001)}]{hale-2001-probabilistic}
John Hale. 2001.
\newblock \href {https://aclanthology.org/N01-1021} {A probabilistic {E}arley parser as a psycholinguistic model}.
\newblock In \emph{Second Meeting of the North {A}merican Chapter of the Association for Computational Linguistics}.

\bibitem[{Hao et~al.(2018)Hao, Merrill, Angluin, Frank, Amsel, Benz, and Mendelsohn}]{hao-etal-2018-context}
Yiding Hao, William Merrill, Dana Angluin, Robert Frank, Noah Amsel, Andrew Benz, and Simon Mendelsohn. 2018.
\newblock \href {https://doi.org/10.18653/v1/W18-5433} {Context-free transductions with neural stacks}.
\newblock In \emph{Proceedings of the 2018 {EMNLP} Workshop {B}lackbox{NLP}: Analyzing and Interpreting Neural Networks for {NLP}}, pages 306--315, Brussels, Belgium. Association for Computational Linguistics.

\bibitem[{Hauser et~al.(2002)Hauser, Chomsky, and Fitch}]{hauser2002faculty}
Marc~D. Hauser, Noam Chomsky, and W.~Tecumseh Fitch. 2002.
\newblock \href {https://doi.org/10.1126/science.298.5598.1569} {The faculty of language: What is it, who has it, and how did it evolve?}
\newblock \emph{Science}, 298(5598):1569--1579.

\bibitem[{Hessel and Schofield(2021)}]{hessel-schofield-2021-effective}
Jack Hessel and Alexandra Schofield. 2021.
\newblock \href {https://doi.org/10.18653/v1/2021.acl-short.27} {How effective is {BERT} without word ordering? implications for language understanding and data privacy}.
\newblock In \emph{Proceedings of the 59th Annual Meeting of the Association for Computational Linguistics and the 11th International Joint Conference on Natural Language Processing (Volume 2: Short Papers)}, pages 204--211, Online. Association for Computational Linguistics.

\bibitem[{Hewitt et~al.(2020)Hewitt, Hahn, Ganguli, Liang, and Manning}]{hewitt-etal-2020-rnns}
John Hewitt, Michael Hahn, Surya Ganguli, Percy Liang, and Christopher~D. Manning. 2020.
\newblock \href {https://doi.org/10.18653/v1/2020.emnlp-main.156} {{RNN}s can generate bounded hierarchical languages with optimal memory}.
\newblock In \emph{Proceedings of the 2020 Conference on Empirical Methods in Natural Language Processing (EMNLP)}, pages 1978--2010, Online. Association for Computational Linguistics.

\bibitem[{Hu et~al.(2020)Hu, Gauthier, Qian, Wilcox, and Levy}]{hu-etal-2020-systematic}
Jennifer Hu, Jon Gauthier, Peng Qian, Ethan Wilcox, and Roger Levy. 2020.
\newblock \href {https://doi.org/10.18653/v1/2020.acl-main.158} {A systematic assessment of syntactic generalization in neural language models}.
\newblock In \emph{Proceedings of the 58th Annual Meeting of the Association for Computational Linguistics}, pages 1725--1744, Online. Association for Computational Linguistics.

\bibitem[{Huang et~al.(2023)Huang, Zelikman, Chen, Wu, Valiant, and Liang}]{huang2023lexinvariant}
Qian Huang, Eric Zelikman, Sarah~Li Chen, Yuhuai Wu, Gregory Valiant, and Percy Liang. 2023.
\newblock \href {http://arxiv.org/abs/2305.16349} {Lexinvariant language models}.

\bibitem[{Jin et~al.(2018)Jin, Doshi-Velez, Miller, Schuler, and Schwartz}]{jin-etal-2018-depth}
Lifeng Jin, Finale Doshi-Velez, Timothy Miller, William Schuler, and Lane Schwartz. 2018.
\newblock \href {https://doi.org/10.18653/v1/D18-1292} {Depth-bounding is effective: Improvements and evaluation of unsupervised {PCFG} induction}.
\newblock In \emph{Proceedings of the 2018 Conference on Empirical Methods in Natural Language Processing}, pages 2721--2731, Brussels, Belgium. Association for Computational Linguistics.

\bibitem[{Joshi(1985)}]{joshi1985TAG}
Aravind~K. Joshi. 1985.
\newblock \href {https://doi.org/10.1017/CBO9780511597855.007} {\emph{Tree adjoining grammars: How much context-sensitivity is required to provide reasonable structural descriptions?}}, Studies in Natural Language Processing, page 206–250. Cambridge University Press.

\bibitem[{Karamcheti et~al.(2021)Karamcheti, Orr, Bolton, Zhang, Goel, Narayan, Bommasani, Narayanan, Hashimoto, Jurafsky, Manning, Potts, Ré, and Liang}]{Mistral}
Siddharth* Karamcheti, Laurel* Orr, Jason Bolton, Tianyi Zhang, Karan Goel, Avanika Narayan, Rishi Bommasani, Deepak Narayanan, Tatsunori Hashimoto, Dan Jurafsky, Christopher~D. Manning, Christopher Potts, Christopher Ré, and Percy Liang. 2021.
\newblock \href {https://github.com/stanford-crfm/mistral} {Mistral - a journey towards reproducible language model training}.

\bibitem[{Karlsson(2007)}]{karlsson2007constraints}
Fred Karlsson. 2007.
\newblock \href {https://doi.org/10.1017/S0022226707004616} {Constraints on multiple center-embedding of clauses}.
\newblock \emph{Journal of Linguistics}, 43(2):365–392.

\bibitem[{Kazemnejad et~al.(2023)Kazemnejad, Padhi, Ramamurthy, Das, and Reddy}]{kazemnejad2023impact}
Amirhossein Kazemnejad, Inkit Padhi, Karthikeyan~Natesan Ramamurthy, Payel Das, and Siva Reddy. 2023.
\newblock The impact of positional encoding on length generalization in transformers.
\newblock \emph{arXiv preprint arXiv:2305.19466}.

\bibitem[{Levy(2008)}]{levy2008expectation}
Roger Levy. 2008.
\newblock \href {https://doi.org/https://doi.org/10.1016/j.cognition.2007.05.006} {Expectation-based syntactic comprehension}.
\newblock \emph{Cognition}, 106(3):1126--1177.

\bibitem[{Mansfield and Kemp(2023)}]{mansfield2023emergence}
John Mansfield and Charles Kemp. 2023.
\newblock The emergence of grammatical structure from inter-predictability.

\bibitem[{Marvin and Linzen(2018)}]{marvin-linzen-2018-targeted}
Rebecca Marvin and Tal Linzen. 2018.
\newblock \href {https://doi.org/10.18653/v1/D18-1151} {Targeted syntactic evaluation of language models}.
\newblock In \emph{Proceedings of the 2018 Conference on Empirical Methods in Natural Language Processing}, pages 1192--1202, Brussels, Belgium. Association for Computational Linguistics.

\bibitem[{Merrill(2019)}]{merrill-2019-sequential}
William Merrill. 2019.
\newblock \href {https://doi.org/10.18653/v1/W19-3901} {Sequential neural networks as automata}.
\newblock In \emph{Proceedings of the Workshop on Deep Learning and Formal Languages: Building Bridges}, pages 1--13, Florence. Association for Computational Linguistics.

\bibitem[{Merrill et~al.(2020)Merrill, Weiss, Goldberg, Schwartz, Smith, and Yahav}]{merrill-etal-2020-formal}
William Merrill, Gail Weiss, Yoav Goldberg, Roy Schwartz, Noah~A. Smith, and Eran Yahav. 2020.
\newblock \href {https://doi.org/10.18653/v1/2020.acl-main.43} {A formal hierarchy of {RNN} architectures}.
\newblock In \emph{Proceedings of the 58th Annual Meeting of the Association for Computational Linguistics}, pages 443--459, Online. Association for Computational Linguistics.

\bibitem[{Mitchell and Bowers(2020)}]{mitchell-bowers-2020-priorless}
Jeff Mitchell and Jeffrey Bowers. 2020.
\newblock \href {https://doi.org/10.18653/v1/2020.coling-main.451} {Priorless recurrent networks learn curiously}.
\newblock In \emph{Proceedings of the 28th International Conference on Computational Linguistics}, pages 5147--5158, Barcelona, Spain (Online). International Committee on Computational Linguistics.

\bibitem[{Moro et~al.(2023)Moro, Greco, and Cappa}]{moro2023impossible}
Andrea Moro, Matteo Greco, and Stefano~F. Cappa. 2023.
\newblock \href {https://doi.org/https://doi.org/10.1016/j.cortex.2023.07.003} {Large languages, impossible languages and human brains}.
\newblock \emph{Cortex}, 167:82--85.

\bibitem[{Murty et~al.(2023)Murty, Sharma, Andreas, and Manning}]{murty2023pushdown}
Shikhar Murty, Pratyusha Sharma, Jacob Andreas, and Christopher~D. Manning. 2023.
\newblock \href {http://arxiv.org/abs/2310.19089} {Pushdown layers: Encoding recursive structure in transformer language models}.

\bibitem[{Musso et~al.(2003)Musso, Moro, Glauche, Rijntjes, Reichenbach, B{\"u}chel, and Weiller}]{musso2003broca}
Mariacristina Musso, Andrea Moro, Volkmar Glauche, Michel Rijntjes, J{\"u}rgen Reichenbach, Christian B{\"u}chel, and Cornelius Weiller. 2003.
\newblock Broca's area and the language instinct.
\newblock \emph{Nature Neuroscience}, 6(7):774--781.

\bibitem[{Nefdt(2024)}]{Nefdt_2024}
Ryan~M. Nefdt. 2024.
\newblock \emph{The Philosophy of Theoretical Linguistics: A Contemporary Outlook}.
\newblock Cambridge University Press.

\bibitem[{Papadimitriou et~al.(2022)Papadimitriou, Futrell, and Mahowald}]{papadimitriou-etal-2022-classifying-grammatical}
Isabel Papadimitriou, Richard Futrell, and Kyle Mahowald. 2022.
\newblock \href {https://doi.org/10.18653/v1/2022.acl-short.71} {When classifying grammatical role, {BERT} doesn{'}t care about word order... except when it matters}.
\newblock In \emph{Proceedings of the 60th Annual Meeting of the Association for Computational Linguistics (Volume 2: Short Papers)}, pages 636--643, Dublin, Ireland. Association for Computational Linguistics.

\bibitem[{Papadimitriou and Jurafsky(2023)}]{papadimitriou2023injecting}
Isabel Papadimitriou and Dan Jurafsky. 2023.
\newblock \href {http://arxiv.org/abs/2304.13060} {Injecting structural hints: Using language models to study inductive biases in language learning}.

\bibitem[{Pham et~al.(2021)Pham, Bui, Mai, and Nguyen}]{pham-etal-2021-order}
Thang Pham, Trung Bui, Long Mai, and Anh Nguyen. 2021.
\newblock \href {https://doi.org/10.18653/v1/2021.findings-acl.98} {Out of order: How important is the sequential order of words in a sentence in natural language understanding tasks?}
\newblock In \emph{Findings of the Association for Computational Linguistics: ACL-IJCNLP 2021}, pages 1145--1160, Online. Association for Computational Linguistics.

\bibitem[{Prasad et~al.(2019)Prasad, van Schijndel, and Linzen}]{prasad-etal-2019-using}
Grusha Prasad, Marten van Schijndel, and Tal Linzen. 2019.
\newblock \href {https://doi.org/10.18653/v1/K19-1007} {Using priming to uncover the organization of syntactic representations in neural language models}.
\newblock In \emph{Proceedings of the 23rd Conference on Computational Natural Language Learning (CoNLL)}, pages 66--76, Hong Kong, China. Association for Computational Linguistics.

\bibitem[{Pérez et~al.(2021)Pérez, Barceló, and Marinkovic}]{perez2021attention}
Jorge Pérez, Pablo Barceló, and Javier Marinkovic. 2021.
\newblock \href {http://jmlr.org/papers/v22/20-302.html} {Attention is {T}uring-complete}.
\newblock \emph{Journal of Machine Learning Research}, 22(75):1--35.

\bibitem[{Qi et~al.(2020)Qi, Zhang, Zhang, Bolton, and Manning}]{qi2020stanza}
Peng Qi, Yuhao Zhang, Yuhui Zhang, Jason Bolton, and Christopher~D. Manning. 2020.
\newblock Stanza: A {Python} natural language processing toolkit for many human languages.
\newblock In \emph{Proceedings of the 58th Annual Meeting of the Association for Computational Linguistics: System Demonstrations}.

\bibitem[{Radford et~al.(2018)Radford, Narasimhan, Salimans, and Sutskever}]{radford2018improving}
Alec Radford, Karthik Narasimhan, Tim Salimans, and Ilya Sutskever. 2018.
\newblock \href {https://openai.com/blog/language-unsupervised/} {Improving language understanding by generative pre-training}.
\newblock Ms, OpenAI.

\bibitem[{Radford et~al.(2019)Radford, Wu, Child, Luan, Amodei, and Sutskever}]{radford2019languagemodels}
Alec Radford, Jeff Wu, Rewon Child, David Luan, Dario Amodei, and Ilya Sutskever. 2019.
\newblock Language models are unsupervised multitask learners.
\newblock Ms, OpenAI.

\bibitem[{Shieber(1985)}]{shieber1985evidence}
Stuart~M. Shieber. 1985.
\newblock Evidence against the context-freeness of natural language.
\newblock \emph{Linguistics and Philosophy}, 8(3):333--343.

\bibitem[{Sinha et~al.(2021)Sinha, Jia, Hupkes, Pineau, Williams, and Kiela}]{sinha-etal-2021-masked}
Koustuv Sinha, Robin Jia, Dieuwke Hupkes, Joelle Pineau, Adina Williams, and Douwe Kiela. 2021.
\newblock \href {https://doi.org/10.18653/v1/2021.emnlp-main.230} {Masked language modeling and the distributional hypothesis: Order word matters pre-training for little}.
\newblock In \emph{Proceedings of the 2021 Conference on Empirical Methods in Natural Language Processing}, pages 2888--2913, Online and Punta Cana, Dominican Republic. Association for Computational Linguistics.

\bibitem[{Tenney et~al.(2019)Tenney, Xia, Chen, Wang, Poliak, McCoy, Kim, Durme, Bowman, Das, and Pavlick}]{tenney2019probing}
Ian Tenney, Patrick Xia, Berlin Chen, Alex Wang, Adam Poliak, R.~Thomas McCoy, Najoung Kim, Benjamin~Van Durme, Samuel~R. Bowman, Dipanjan Das, and Ellie Pavlick. 2019.
\newblock \href {https://openreview.net/forum?id=SJzSgnRcKX} {What do you learn from context? {Probing} for sentence structure in contextualized word representations}.
\newblock In \emph{International Conference on Learning Representations}.

\bibitem[{Vaswani et~al.(2017)Vaswani, Shazeer, Parmar, Uszkoreit, Jones, Gomez, Kaiser, and Polosukhin}]{vaswani2017attention}
Ashish Vaswani, Noam Shazeer, Niki Parmar, Jakob Uszkoreit, Llion Jones, Aidan~N Gomez, \L~ukasz Kaiser, and Illia Polosukhin. 2017.
\newblock \href {https://proceedings.neurips.cc/paper_files/paper/2017/file/3f5ee243547dee91fbd053c1c4a845aa-Paper.pdf} {Attention is all you need}.
\newblock In \emph{Advances in Neural Information Processing Systems}, volume~30. Curran Associates, Inc.

\bibitem[{Warstadt et~al.(2023)Warstadt, Choshen, Mueller, Williams, Wilcox, and Zhuang}]{warstadt2023papers}
Alex Warstadt, Leshem Choshen, Aaron Mueller, Adina Williams, Ethan Wilcox, and Chengxu Zhuang. 2023.
\newblock \href {http://arxiv.org/abs/2301.11796} {Call for papers -- the {BabyLM} challenge: Sample-efficient pretraining on a developmentally plausible corpus}.

\bibitem[{Weiss et~al.(2018)Weiss, Goldberg, and Yahav}]{weiss-etal-2018-practical}
Gail Weiss, Yoav Goldberg, and Eran Yahav. 2018.
\newblock \href {https://doi.org/10.18653/v1/P18-2117} {On the practical computational power of finite precision {RNN}s for language recognition}.
\newblock In \emph{Proceedings of the 56th Annual Meeting of the Association for Computational Linguistics (Volume 2: Short Papers)}, pages 740--745, Melbourne, Australia. Association for Computational Linguistics.

\bibitem[{Wilcox et~al.(2018)Wilcox, Levy, Morita, and Futrell}]{wilcox-etal-2018-rnn}
Ethan Wilcox, Roger Levy, Takashi Morita, and Richard Futrell. 2018.
\newblock \href {https://doi.org/10.18653/v1/W18-5423} {What do {RNN} language models learn about filler{--}gap dependencies?}
\newblock In \emph{Proceedings of the 2018 {EMNLP} Workshop {B}lackbox{NLP}: Analyzing and Interpreting Neural Networks for {NLP}}, pages 211--221, Brussels, Belgium. Association for Computational Linguistics.

\bibitem[{Wilcox et~al.(2023)Wilcox, Futrell, and Levy}]{wilcox2023syntactic}
Ethan~Gotlieb Wilcox, Richard Futrell, and Roger Levy. 2023.
\newblock \href {https://doi.org/10.1162/ling_a_00491} {Using computational models to test syntactic learnability}.
\newblock \emph{Linguistic Inquiry}, pages 1--44.

\bibitem[{Wu et~al.(2023{\natexlab{a}})Wu, D'Oosterlinck, Geiger, Zur, and Potts}]{pmlr-v202-wu23b}
Zhengxuan Wu, Karel D'Oosterlinck, Atticus Geiger, Amir Zur, and Christopher Potts. 2023{\natexlab{a}}.
\newblock \href {https://proceedings.mlr.press/v202/wu23b.html} {Causal proxy models for concept-based model explanations}.
\newblock In \emph{Proceedings of the 40th International Conference on Machine Learning}, volume 202 of \emph{Proceedings of Machine Learning Research}, pages 37313--37334. PMLR.

\bibitem[{Wu et~al.(2023{\natexlab{b}})Wu, Geiger, Icard, Potts, and Goodman}]{wu-etal-2023-Boundless-DAS}
Zhengxuan Wu, Atticus Geiger, Thomas Icard, Christopher Potts, and Noah Goodman. 2023{\natexlab{b}}.
\newblock \href {https://proceedings.neurips.cc/paper_files/paper/2023/file/f6a8b109d4d4fd64c75e94aaf85d9697-Paper-Conference.pdf} {Interpretability at scale: Identifying causal mechanisms in {Alpaca}}.
\newblock In \emph{Advances in Neural Information Processing Systems}, volume~36, pages 78205--78226. Curran Associates, Inc.

\bibitem[{Wu et~al.(2022)Wu, Geiger, Rozner, Kreiss, Lu, Icard, Potts, and Goodman}]{wu-etal-2022-causal}
Zhengxuan Wu, Atticus Geiger, Joshua Rozner, Elisa Kreiss, Hanson Lu, Thomas Icard, Christopher Potts, and Noah Goodman. 2022.
\newblock \href {https://doi.org/10.18653/v1/2022.naacl-main.318} {Causal distillation for language models}.
\newblock In \emph{Proceedings of the 2022 Conference of the North American Chapter of the Association for Computational Linguistics: Human Language Technologies}, pages 4288--4295, Seattle, United States. Association for Computational Linguistics.

\end{thebibliography}
\bibliographystyle{acl_natbib}

\newpage

\appendix

\section*{Supplementary Materials}

\section{Dataset Filters} \label{sec:appendix-dataset-filters}

The BabyLM dataset \cite{warstadt2023papers} is an English-language dataset of 
about 100 million words intended to approximate the 
amount of linguistic data available to an English-speaking child. 
To create a dataset for an impossible language, we first pre-process the BabyLM dataset using Stanza \cite{qi2020stanza}. We perform sentence segmentation on each dataset file and then extract part-of-speech (POS) and morphological feature tags for all the sentences, which are required for the \textsc{*Hop} transformations. We transform each tagged sentence in the original BabyLM dataset using the impossible language's rule-based perturbation function, as described in \Cref{sec:impossible-languages}. 
Depending on the class of the impossible language and the specific features of the input sentence, perturbed sentences may be included or excluded from the final dataset used for model training (see below %
for details on this filtering). Since we apply these filters, the language classes have datasets of slightly different sizes. The \textsc{*Shuffle} and \textsc{*Reverse} languages have training sets of about 9.69 million sentences, and the \textsc{*Hop} languages have training sets of about 8.43 million sentences.

\paragraph{\textsc{*Shuffle Filters}}
For the \textsc{*Shuffle} languages, we filter sentences from the BabyLM dataset such that the set of token sequence lengths seen in the validation and test sets are also seen in the training set. This ensures that any shuffles for the \textsc{DeterministicShuffle} perturbation (which are determined by the token sequence length) in the test set have also occurred at least once in the training set. We apply these filters for all \textsc{*Shuffle} languages such that their datasets are comprised of the same subset of original sentences.

\paragraph{\textsc{*Reverse Filters}}
For the \textsc{*Reverse} languages, we do not apply any sentence filtering, so their models are trained on the entire BabyLM dataset.

\paragraph{\textsc{*Hop Filters}}

For the \textsc{*Hop} languages, we filter out sentences from the BabyLM dataset that would not allow the special markers to fully complete 4 hops in the \textsc{TokenHop} or \textsc{WordHop} perturbations, i.e.\ sentences in which a 3rd-person present tense verb is too close to the end of the sentence. We again filter out these sentences from all perturbations, so \textsc{TokenHop}, \textsc{WordHop}, and \textsc{NoHop} are comprised of the same subset of original sentences from the BabyLM dataset.

\section{GPT-2 Training Details and Hyperparameters} \label{sec:appendix-hyperparameters}

We train GPT-2 small models with a standard training regime \cite{radford2018improving, radford2019languagemodels} using the library of  \citet{Mistral}. We mostly use the default GPT-2 small hyperparameters to train our models (context length of 1024, batch size of 512, etc.). We only change the total number of training steps and the number of warm-up steps. We train with a learning rate that linearly warms up from 0 to 6e-4 over 300 steps. While 10\% of steps for warm-up is typical for LLM training, we acknowledge that the best warm-up may be different when using a small pretraining dataset, so we also tried 1,000 warm-up steps and 4,000 warm-up steps. (4,000 steps is the GPT-2 default. Since we only train for 3,000 steps, this effectively means we have a learning-rate that linearly warms up from 0 to \mbox{4.5e-4}.) Using a different warm-up did not change the ranking of impossible language model perplexities.

We train the models for 3,000 training steps, which equates to about 11.03 epochs for the \textsc{*Shuffle} languages, 10.05 epochs for the \textsc{*Reverse} languages, and 12.04 epochs for the \textsc{*Hop} languages. The vocabulary set also varies based on the language. The \textsc{*Shuffle} languages use the standard GPT-2 vocabulary containing 50,257 tokens; the \textsc{*Reverse} languages add one special token \revmarker, for a  vocabulary size of 50,258; and the \textsc{*Hop} languages add two special tokens \singularmarker\space and \pluralmarker\space for verb inflection, for a vocabulary size of 50,259. We train on NVIDIA RTX 3090 (24GB) GPUs and NVIDIA RTX A6000 (48GB) GPUs. The runtime for each pretraining experiment was $\sim$24 hours (for one language and one random seed), for a total experiment runtime of $\sim$1800 hours.

\section{Results for Models without Positional Encodings} \label{sec:appendix-no-pos-encoding}

\begin{figure*}[ht]
    \centering
    \centering
    \includegraphics[width=0.95\textwidth]{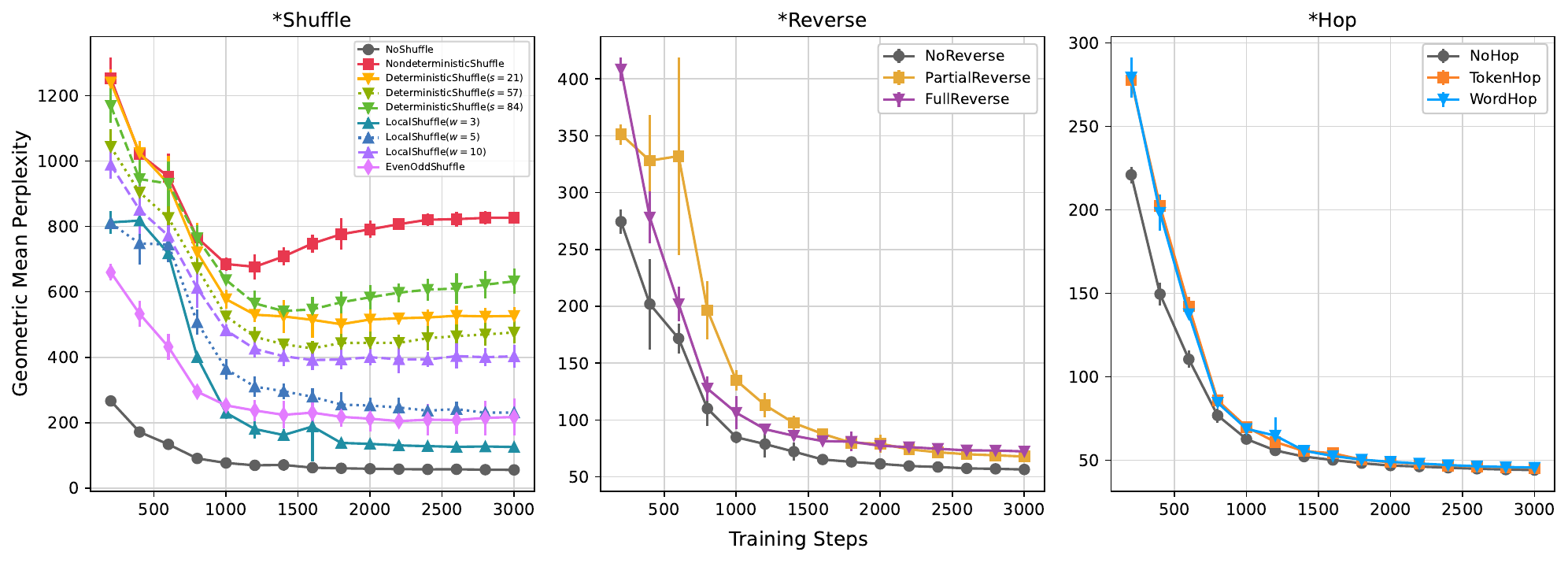}
    \caption{Perplexities on a sample of 10K test sentences for each
    impossible language model trained \emph{without positional encodings}. Error bars indicate 95\% confidence intervals across 5 training runs initialized with different random seeds and evaluated on different test samples.}
    \label{fig:perplexities-no-pos-encodings}
    
    \bigskip
    \centering
    \begin{subfigure}[t]{0.49\textwidth}
        \centering
        \includegraphics[width=0.95\linewidth]{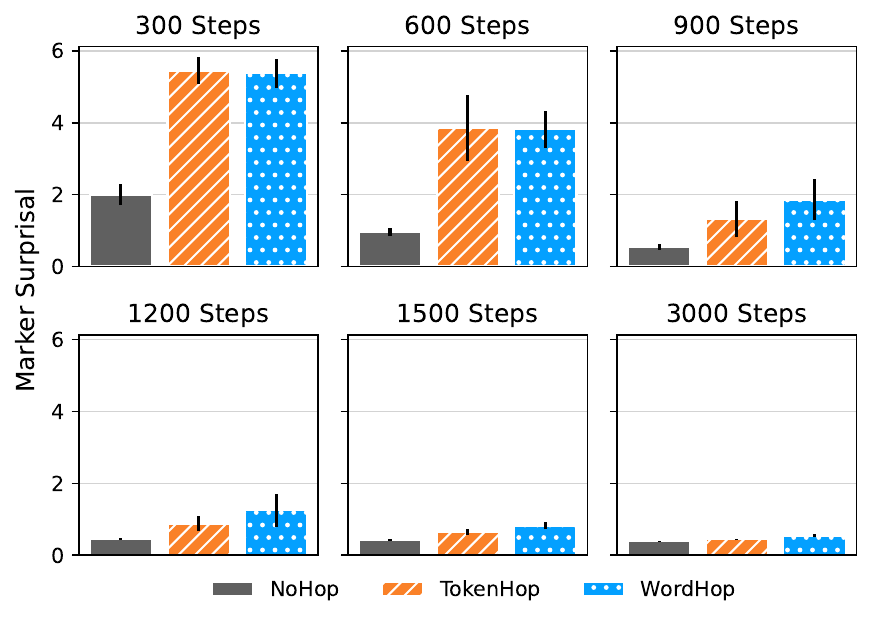}
        \caption{Mean surprisals of the verb marker token (\singularmarker
        \space or \pluralmarker) for each \textsc{*Hop} model. }
    \end{subfigure}\hfill
    \begin{subfigure}[t]{0.49\textwidth}
        \centering
        \includegraphics[width=0.95\linewidth]{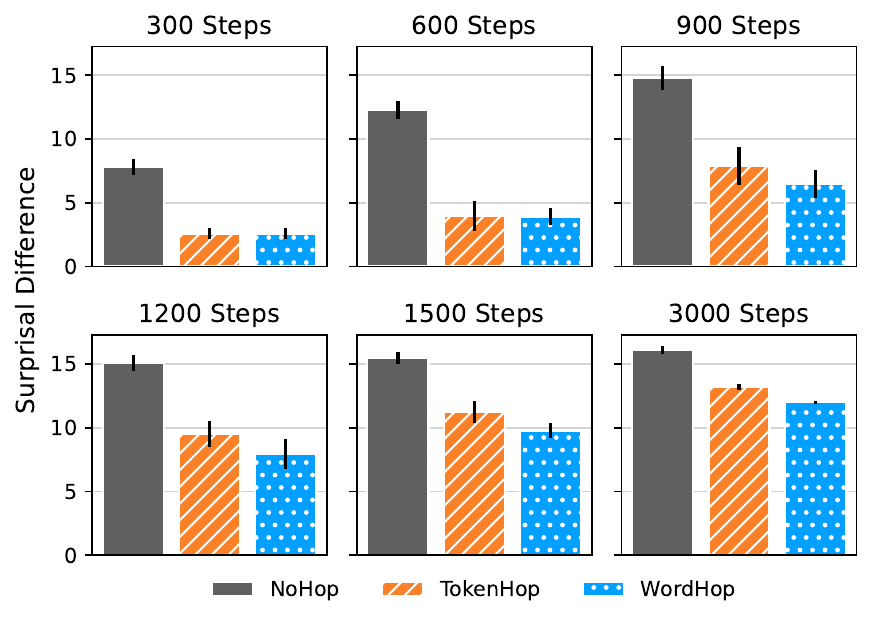}
        \caption{Mean surprisal difference between the verb marker token (\singularmarker
        \space or \pluralmarker) and the following token for each \textsc{*Hop} model.}
    \end{subfigure}
    \caption{Surprisal tests for each \textsc{*Hop} model over training steps (trained \emph{without positional encodings}). Error bars indicate 95\% confidence intervals across 5 training runs initialized with different random seeds and evaluated on different test samples.}
    \label{fig:surprisals_no_pos_encodings}
    
    \bigskip
    \centering
    \includegraphics[width=0.329\textwidth]{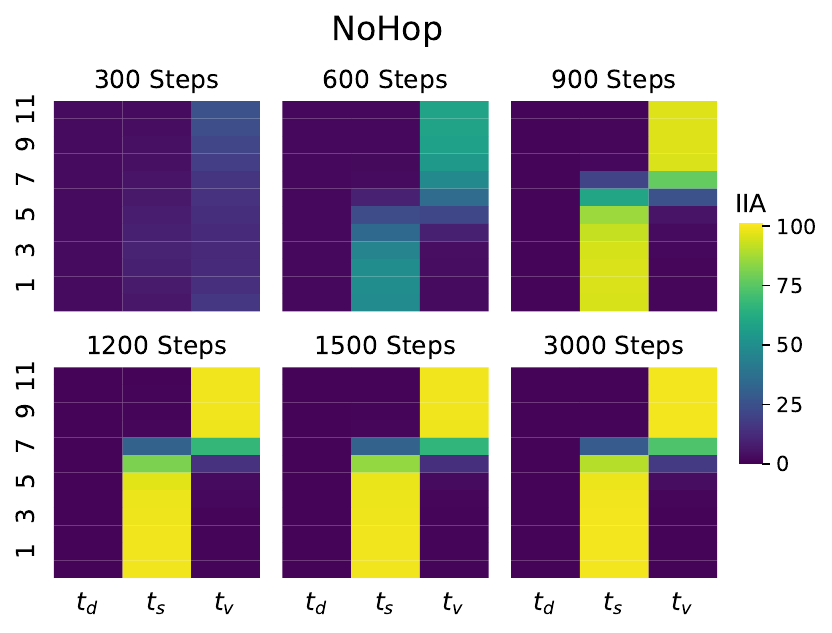}
    \includegraphics[width=0.329\textwidth]{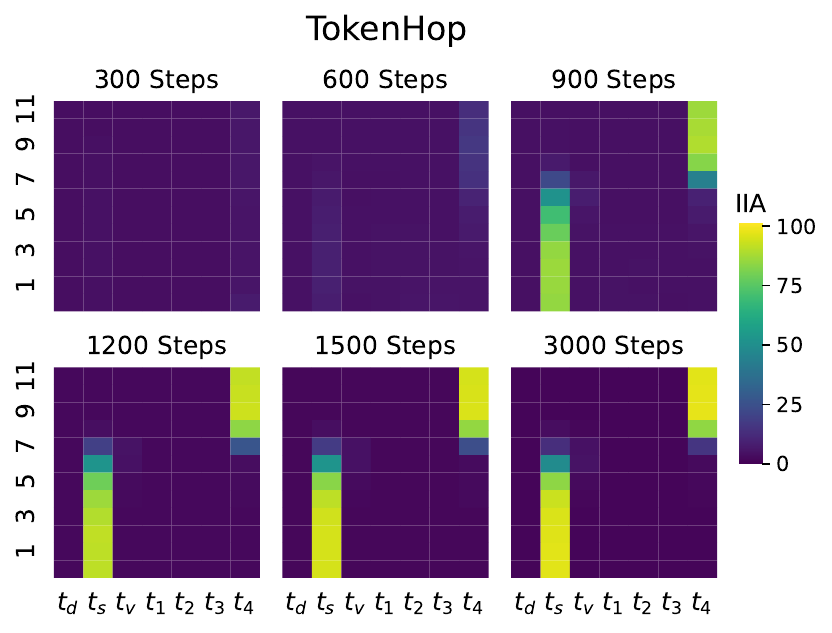}
    \includegraphics[width=0.329\textwidth]{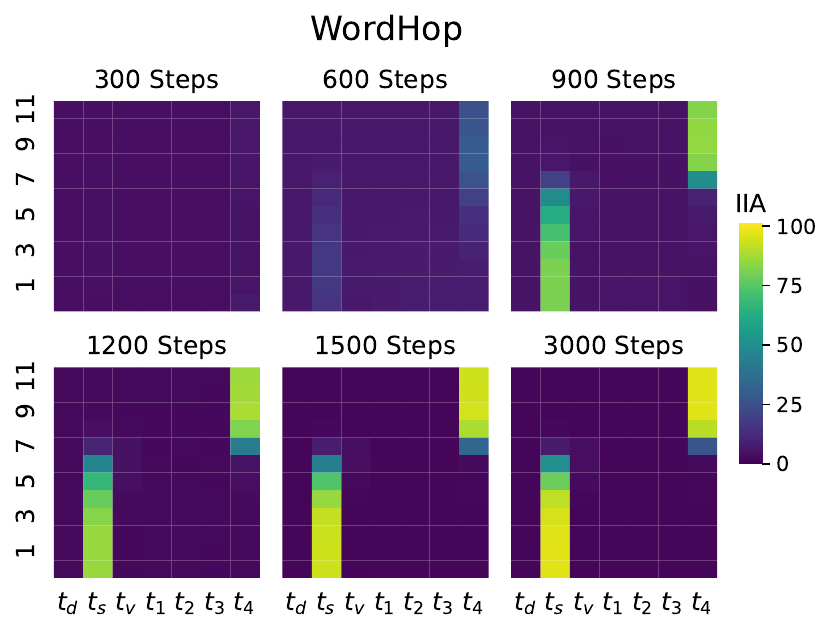}
    \caption{Subject--verb agreement interchange intervention accuracies (IIA) for each
    \textsc{*Hop} model trained \emph{without positional encodings}. Vertical axes denote the GPT-2 layer of the intervention, and horizontal axes denote the token position of the intervention.
    $t_d$, $t_s$, and $t_v$ represent the tokens for the determiner, subject, and
    verb, respectively. $t_1 \dots t_4$ represent the four tokens/words between the 
    verb and its marker for \textsc{TokenHop} and \textsc{WordHop}.
    IIA values are averaged over results from 5 models initialized on different random seeds. See Figures~\ref{fig:iia_ci_hop_control_no_pos_encodings},~\ref{fig:iia_ci_hop_tokens4_no_pos_encodings}, and~\ref{fig:iia_ci_hop_words4_no_pos_encodings} for confidence intervals.
    }
    \label{fig:iia_hop_models_no_pos_encodings}
\end{figure*}

Here, we present results for each of our experiments using GPT-2 models we trained without positional encodings. All other aspects of the experiments are the same, including the impossible language datasets and training hyperparameters. We again train 5 sets of models initialized using different random seeds. \Cref{fig:perplexities-no-pos-encodings} presents the perplexity results; \Cref{fig:surprisals_no_pos_encodings} presents the surprisal results; and \Cref{fig:iia_hop_models_no_pos_encodings} presents the causal intervention results.

\section{Constituency Probing Evaluation} \label{sec:appendix-probing}
We also test how perturbations might influence latent linguistic properties in sentences that are seemingly \emph{unaffected} by the perturbations. For this, we develop a constituency probing experiment to examine whether the contextual representations generated by different models are effective in classifying a sequence of tokens with an appropriate constituent label, similar to the edge probing experiments of \citealt{tenney2019probing}. For example, if the input sentence is ``I enjoy strawberry ice cream'' and the span of tokens in question represents the constituent ``strawberry ice cream,'' the span should be labeled as a noun phrase (NP).

\paragraph{Setup.}
We conduct these experiments for \textsc{*Reverse} and \textsc{*Hop} languages, since these languages have constituents in contiguous token sequences. For \textsc{NoReverse} and \textsc{PartialReverse}, we take a sample of unaltered BabyLM test sentences and omit the reversal token \revmarker. For  \textsc{FullReverse}, we use the same sample sentences, but reverse the tokens. For the \textsc{*Hop} languages, we use a sample of BabyLM test sentences that are unaffected by the perturbation, which are sentences that do not contain 3rd-person present tense verbs. To extract constituents for testing, we parse the sample sentences using Stanza's BERT-based consituency parser. We include noun phrases (NP), verb phrases (VP), adjective phrases (ADJP), adverb phrases (ADVP), and prepositional phrases (PP), and we stratify the samples so that there are equal numbers of example constituents for each phrasal category. We obtain a total of 10K examples for probe training and testing for each language class, where an example is comprised of a tokenized sentence, indices of the constituent span, and the constituent label.

Our probes are L2-regularized logistic regression classifiers trained on the span representations of the tokens corresponding to constituents in the examples. To obtain span representations for training the probes, we mean-pool the representations of the tokens within the span. We try extracting representations from GPT-2 by averaging the last four hidden layers of the model or using different layers individually. We train each probe for a maximum of 10 iterations and hold out 20\% of constituent examples for testing.

\paragraph{Hypothesis.}
Constituency probes will achieve higher accuracy for possible languages than impossible ones, in virtue of the fact that the impossible languages are defined by some rules that do not respect constituency boundaries. 

\paragraph{Results.}

\begin{figure*}
    \centering
    \begin{subfigure}[t]{0.45\textwidth}
        \includegraphics[width=1\textwidth]{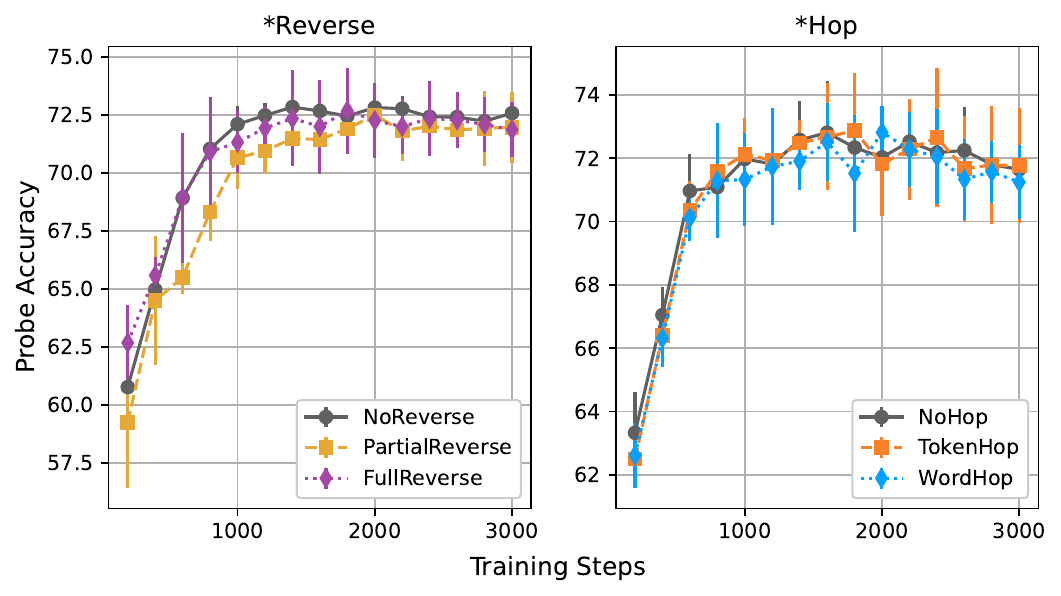}
        \caption{Probe accuracy for \textsc{*Reverse} and \textsc{*Hop} models.}
    \end{subfigure}
    \hspace{1em}
    \begin{subfigure}[t]{0.45\textwidth}
        \includegraphics[width=1\textwidth]{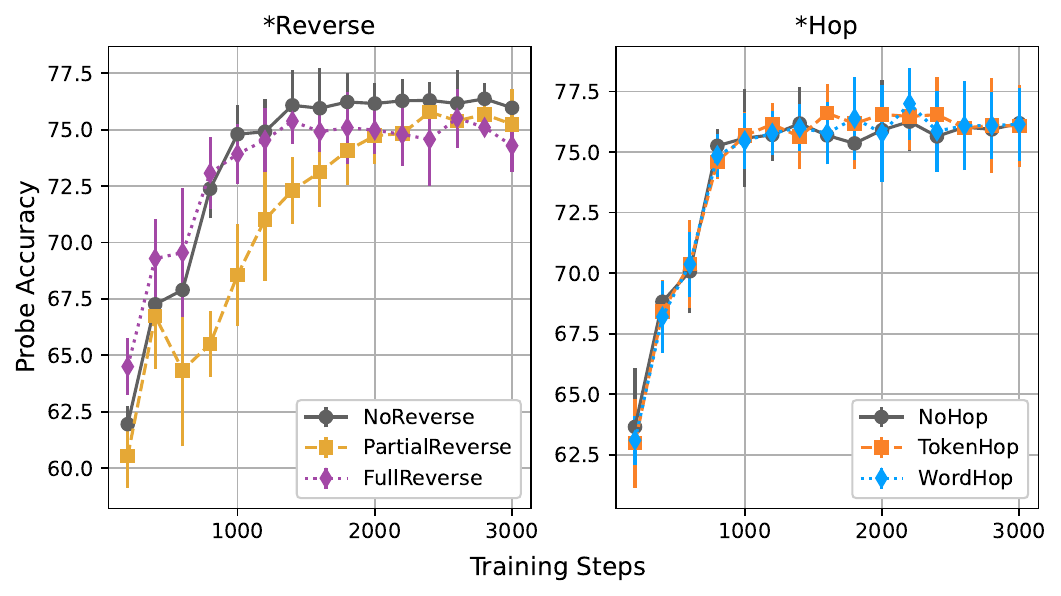}
        \caption{Probe accuracy \emph{without positional encodings}. }
    \end{subfigure}         
    \caption{Constituency probe accuracy for \textsc{*Reverse} and \textsc{*Hop} models over training steps. Span representations were extracted by averaging the last four hidden layers of GPT-2. Error bars indicate 95\% confidence intervals across 5 training runs initialized with different random seeds and evaluated on different test samples.}
    \label{fig:probing}
\end{figure*}

\begin{figure*}
    \centering
     \begin{subfigure}{1\textwidth}
         \centering
         \includegraphics[width=0.9\textwidth]{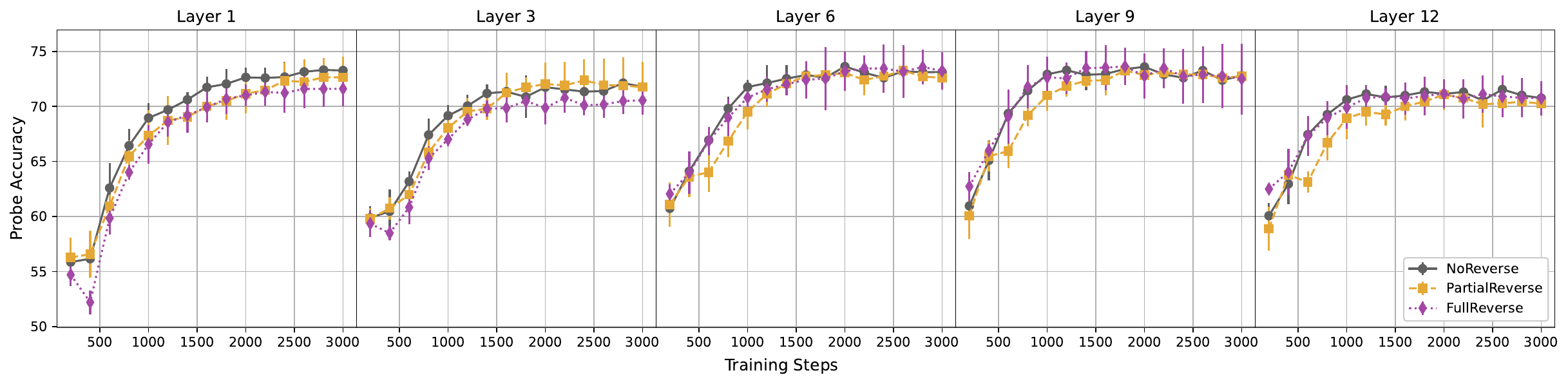}
         \caption{Probe accuracy for \textsc{*Reverse} models.}
     \end{subfigure}
     \begin{subfigure}{1\textwidth}
         \centering
         \includegraphics[width=0.9\textwidth]{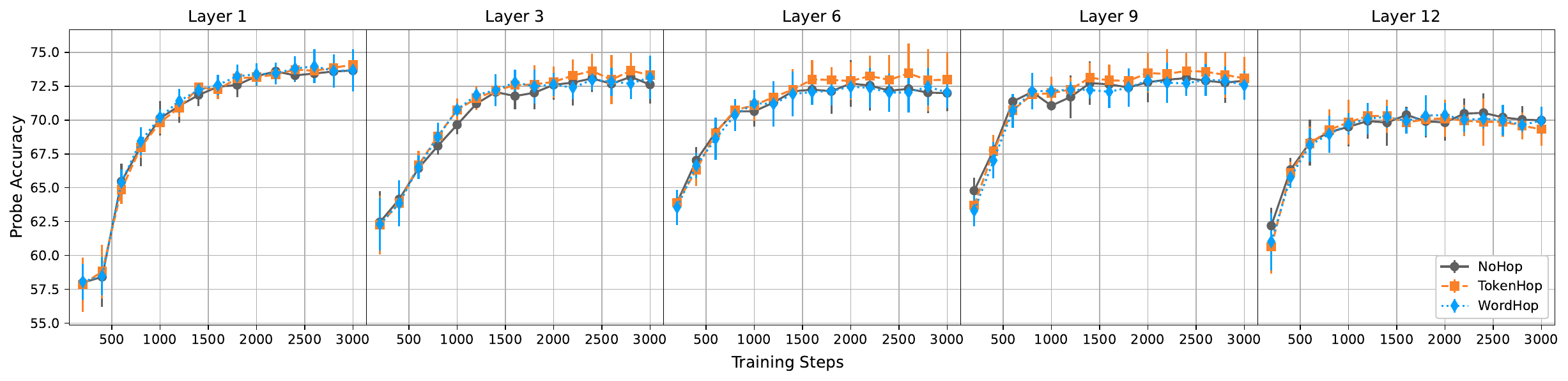}
         \caption{Probe accuracy for \textsc{*Hop} models.}
     \end{subfigure}
     \begin{subfigure}{1\textwidth}
         \centering
         \includegraphics[width=0.9\textwidth]{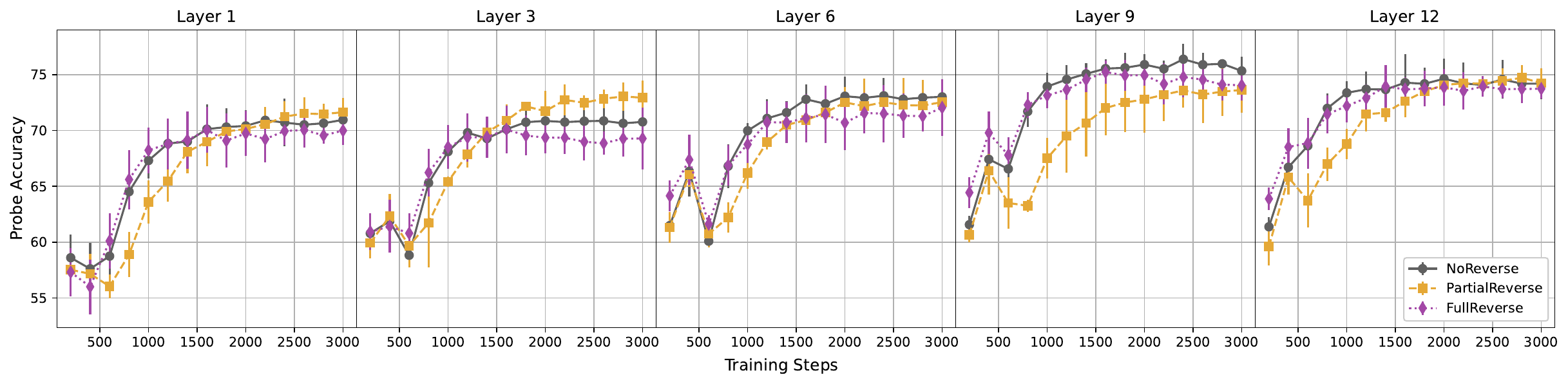}
         \caption{Probe accuracy for \textsc{*Reverse} models \emph{without positional encodings}.}
     \end{subfigure}
     \begin{subfigure}{1\textwidth}
         \centering
         \includegraphics[width=0.9\textwidth]{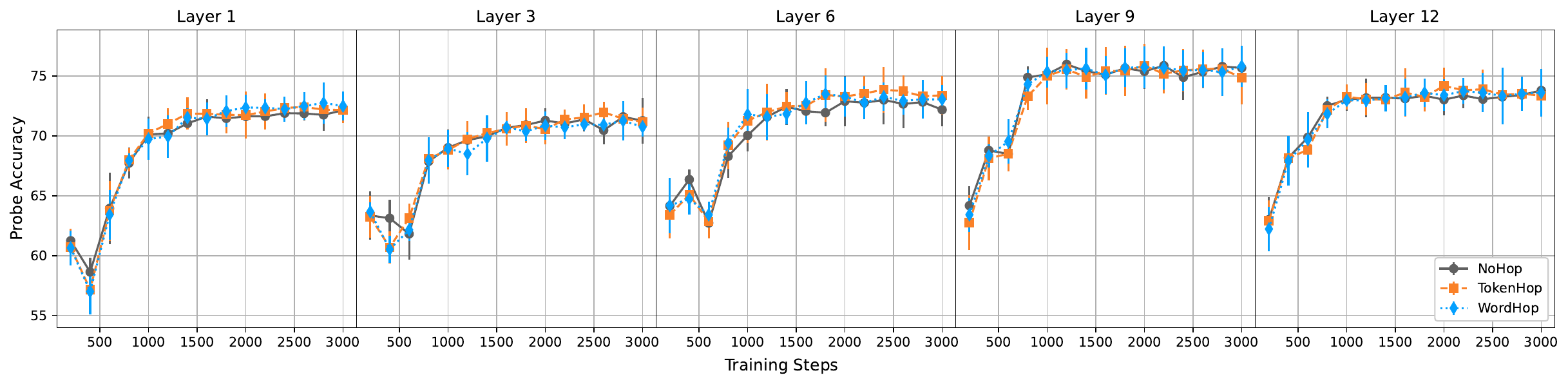}
         \caption{Probe accuracy for \textsc{*Hop} models \emph{without positional encodings}.}
     \end{subfigure}
    \caption{Constituency probe accuracy for \textsc{*Reverse} and \textsc{*Hop} models using span representations extracted from different GPT-2 layers (1, 3, 6, 9, 12) over training steps. Error bars indicate 95\% confidence intervals across 5 training runs initialized with different random seeds and evaluated on different test samples.}
    \label{fig:probing_by_layer}
\end{figure*}

The results of the probing experiment using the average of the last four GPT-2 layers are presented in \Cref{fig:probing}. Across \textsc{*Reverse} and \textsc{*Hop} models trained \emph{with} positional encodings, there are not any clear trends indicating that certain models have better representations of constituents than others, as differences among probe accuracies are minimal and unstable across training steps. However, looking closely at the \textsc{*Reverse} models \emph{without} positional encodings, we can see that \textsc{PartialReverse} has significantly lower probe accuracy than the other models up until 2K training steps. We found similar results when using different layers for span representations, as shown in \Cref{fig:probing_by_layer}. These results might indicate that the \textsc{*Hop} perturbations were too weak to fundamentally affect the models' representations of latent linguistic structure, but quite unnatural reversal rule of the \textsc{PartialReverse} language disturbed consituency boundaries in a way that could not be recovered by GPT-2 models without positional encodings.

\section{Additional \textsc{DeterministicShuffle} Results} \label{sec:appendix-deterministic-shuffle}

\begin{figure*}[ht]
    \centering
     \begin{subfigure}{1\textwidth}
         \centering
         \includegraphics[width=1\textwidth]{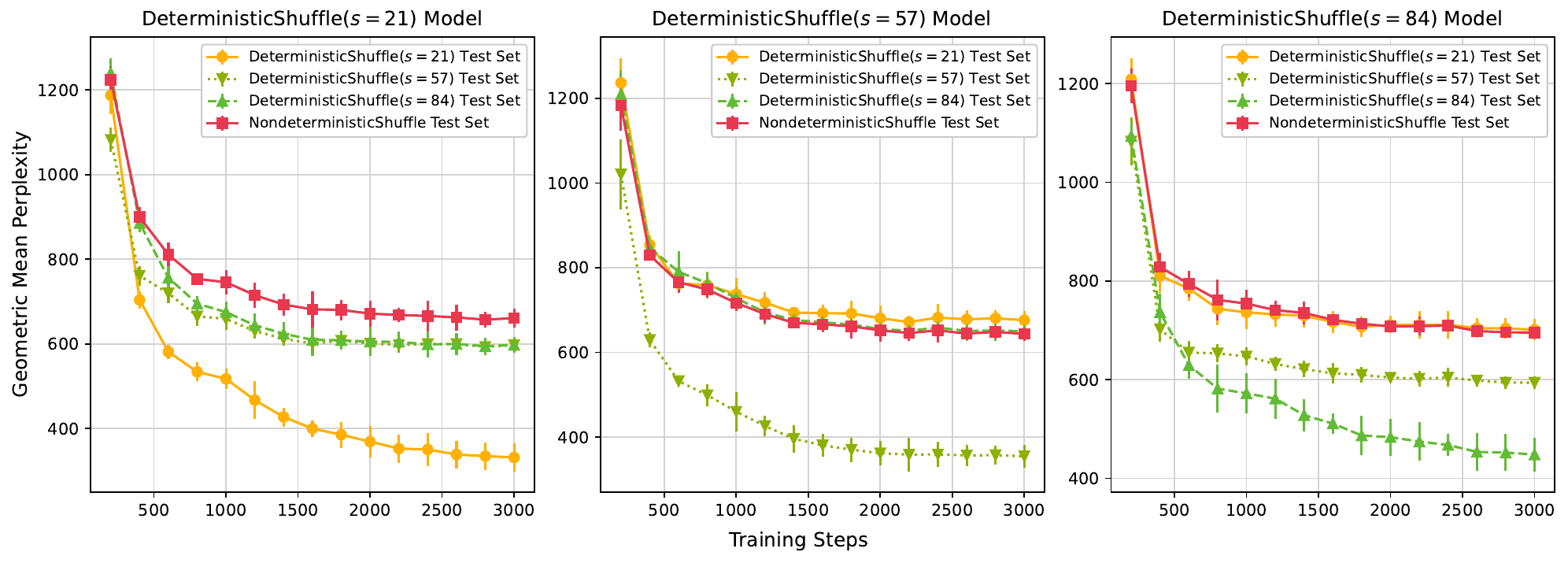}
         \caption{Test perplexities for models \emph{with} positional encodings.}
     \end{subfigure}
     
     \begin{subfigure}{1\textwidth}
         \centering
         \includegraphics[width=1\textwidth]{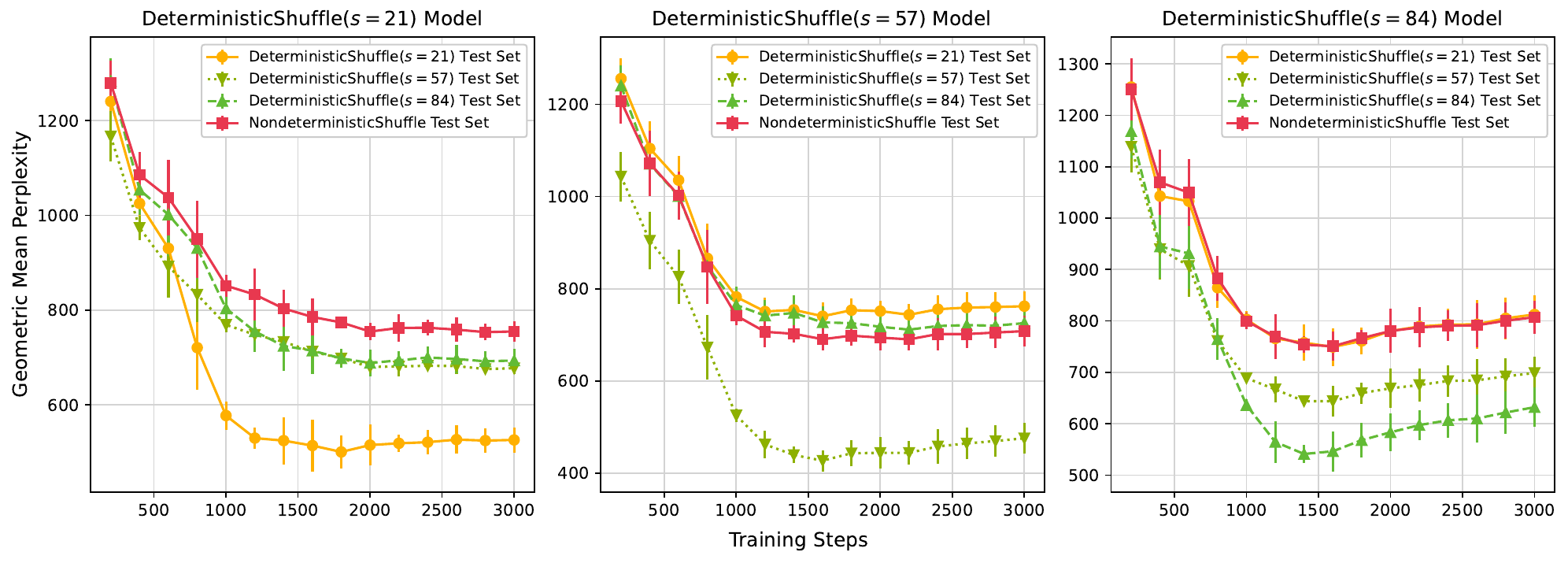}
         \caption{Test perplexities for models \emph{without} positional encodings.}
     \end{subfigure}
    \caption{Test perplexities for each \textsc{DeterministicShuffle} model ($s=21$ left, $s=57$ middle, $s=84$ right) on the $\textsc{NondeterministicShuffle}$ test sample and all other $\textsc{DeterministicShuffle}$ test samples. Perplexities were taken on a sample of 10K test sentences from each shuffled test set. Error bars indicate 95\% confidence intervals across 5 training runs initialized with different random seeds and evaluated on different test samples.}
    \label{fig:perplexities-deterministic-shuffle}
\end{figure*}

In addition to perplexities of each impossible language model on its own test data, we also obtain perplexities for each $\textsc{DeterministicShuffle}$ model on the $\textsc{NondeterministicShuffle}$ test sample and all other $\textsc{DeterministicShuffle}$ test samples. This measures whether these models have learned to distinguish their own shuffles from other shuffles. We found that this was indeed the case, as shown in the results in \Cref{fig:perplexities-deterministic-shuffle}.

\section{Confidence Intervals for Interchange Intervention Accuracies} \label{sec:intervention_confidence_intervals}

\begin{figure*}
    \centering
     \begin{subfigure}{0.3\textwidth}
         \centering
         \includegraphics[width=1\textwidth]{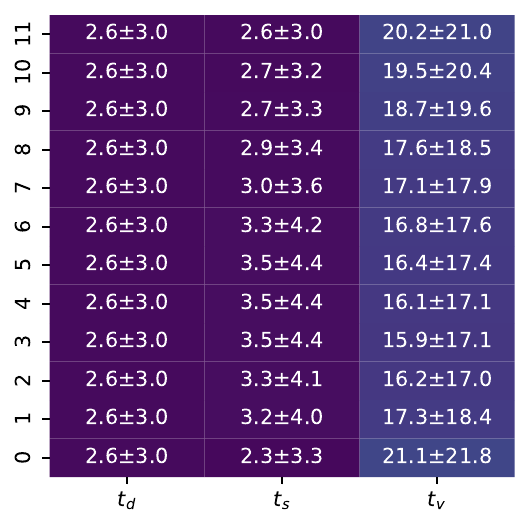}
         \caption{300 Training Steps.}
     \end{subfigure}
     \begin{subfigure}{0.3\textwidth}
         \centering
         \includegraphics[width=1\textwidth]{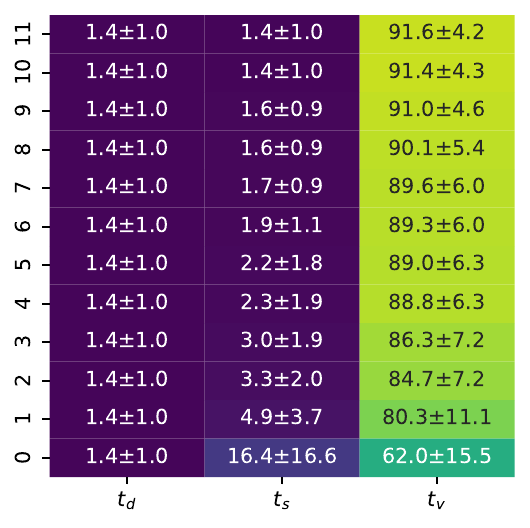}
         \caption{600 Training Steps.}
     \end{subfigure}
     \begin{subfigure}{0.3\textwidth}
         \centering
         \includegraphics[width=1\textwidth]{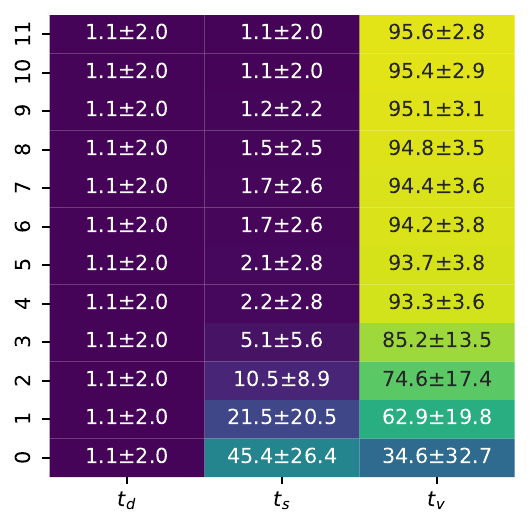}
         \caption{900 Training Steps.}
     \end{subfigure}
     \begin{subfigure}{0.3\textwidth}
         \centering
         \includegraphics[width=1\textwidth]{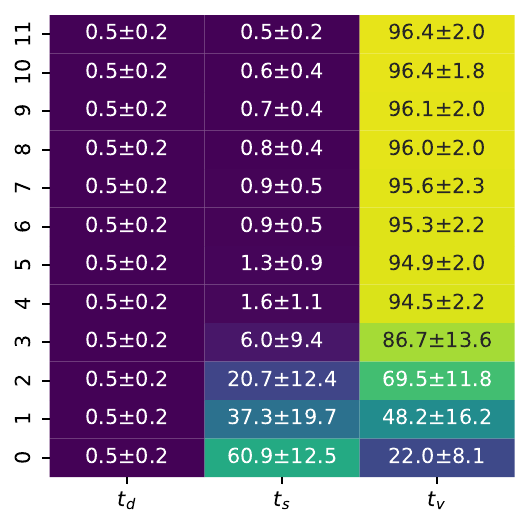}
         \caption{1200 Training Steps.}
     \end{subfigure}
     \begin{subfigure}{0.3\textwidth}
         \centering
         \includegraphics[width=1\textwidth]{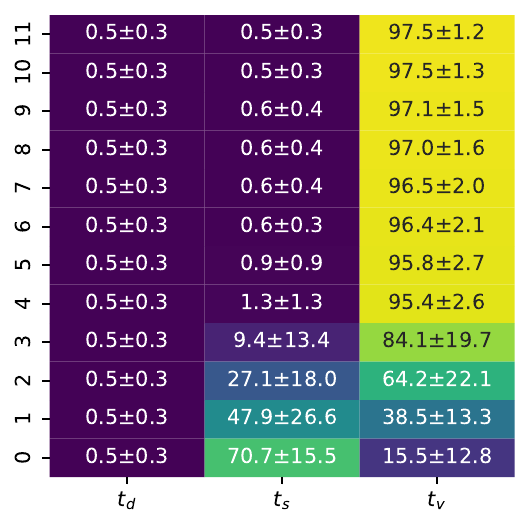}
         \caption{1500 Training Steps.}
     \end{subfigure}
     \begin{subfigure}{0.3\textwidth}
         \centering
         \includegraphics[width=1\textwidth]{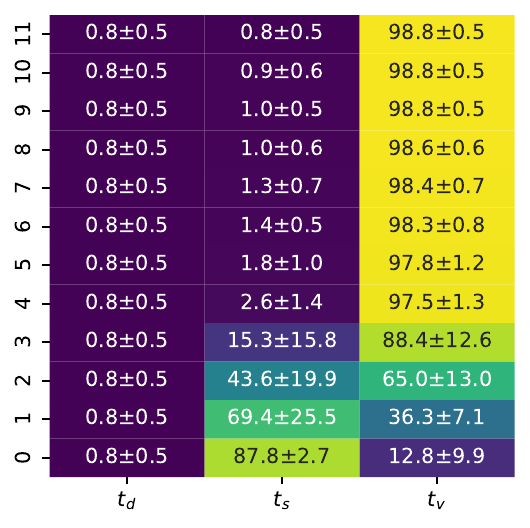}
         \caption{3000 Training Steps.}
     \end{subfigure}
    \caption{Subject--verb agreement interchange intervention accuracies (IIA) for
    \textsc{NoHop}, with confidence intervals across models trained on 5 different random seeds. Vertical axes denote the GPT-2 layer of the intervention, and horizontal axes denote the token position of the intervention. $t_d$, $t_s$, and $t_v$ represent the tokens for the determiner, subject, and verb, respectively.}
    \label{fig:iia_ci_hop_control}
\end{figure*}

\begin{figure*}
    \centering
     \begin{subfigure}{0.49\textwidth}
         \centering
         \includegraphics[width=1\textwidth]{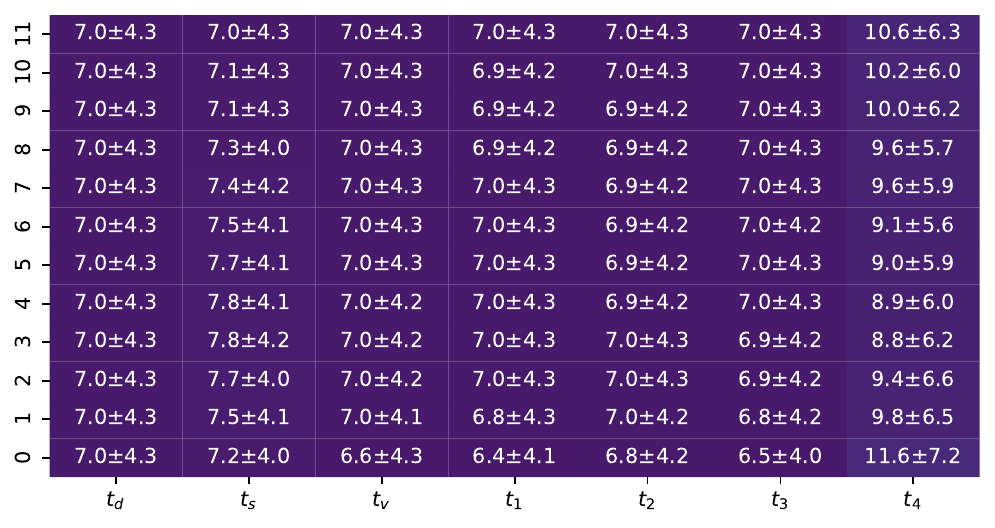}
         \caption{300 Training Steps.}
     \end{subfigure}
     \begin{subfigure}{0.49\textwidth}
         \centering
         \includegraphics[width=1\textwidth]{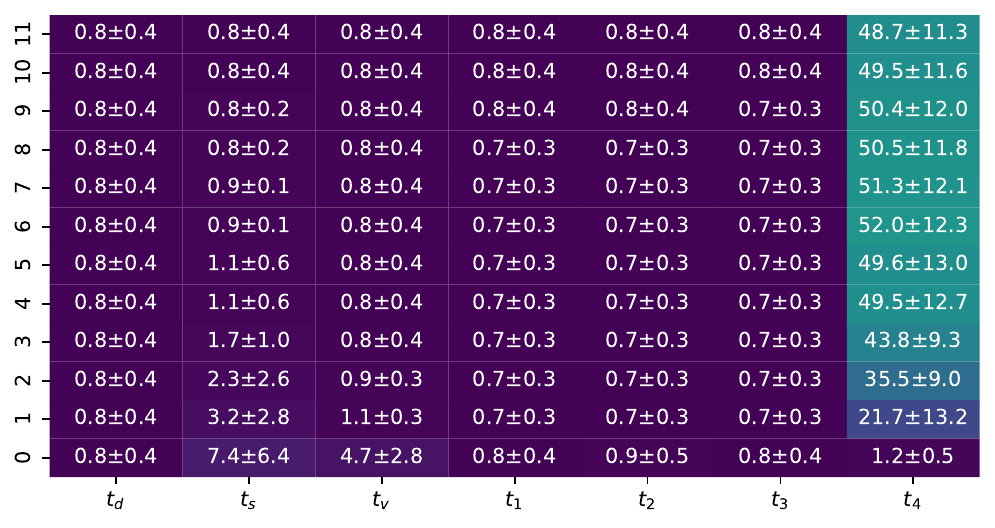}
         \caption{600 Training Steps.}
     \end{subfigure}
     \begin{subfigure}{0.49\textwidth}
         \centering
         \includegraphics[width=1\textwidth]{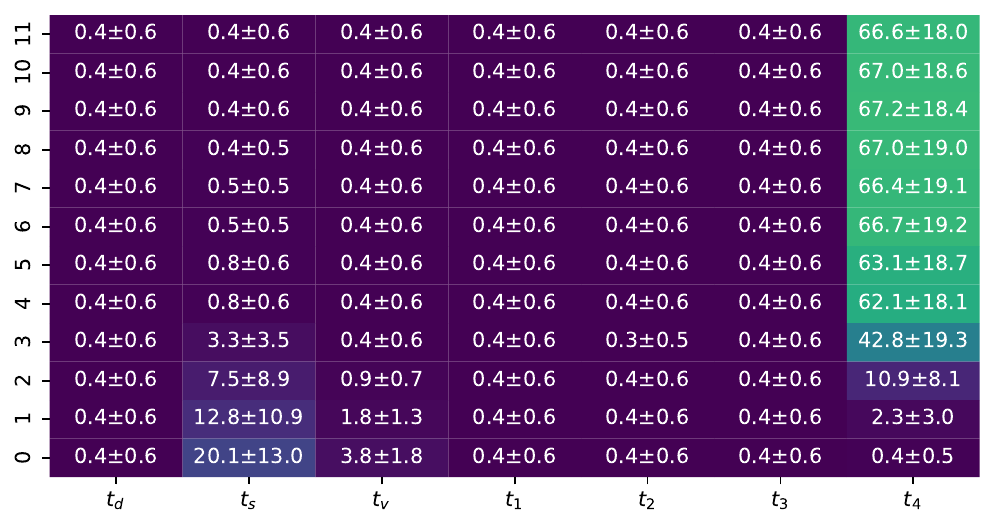}
         \caption{900 Training Steps.}
     \end{subfigure}
     \begin{subfigure}{0.49\textwidth}
         \centering
         \includegraphics[width=1\textwidth]{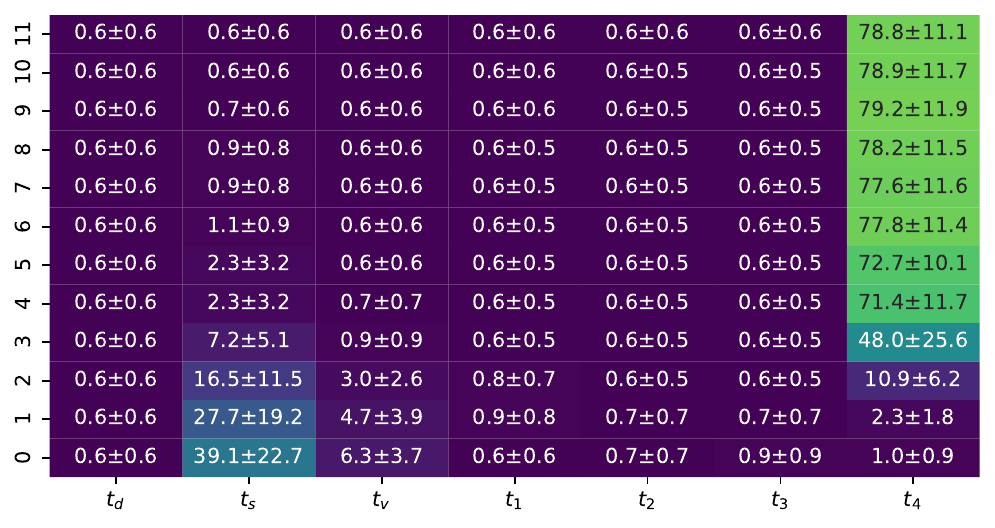}
         \caption{1200 Training Steps.}
     \end{subfigure}
     \begin{subfigure}{0.49\textwidth}
         \centering
         \includegraphics[width=1\textwidth]{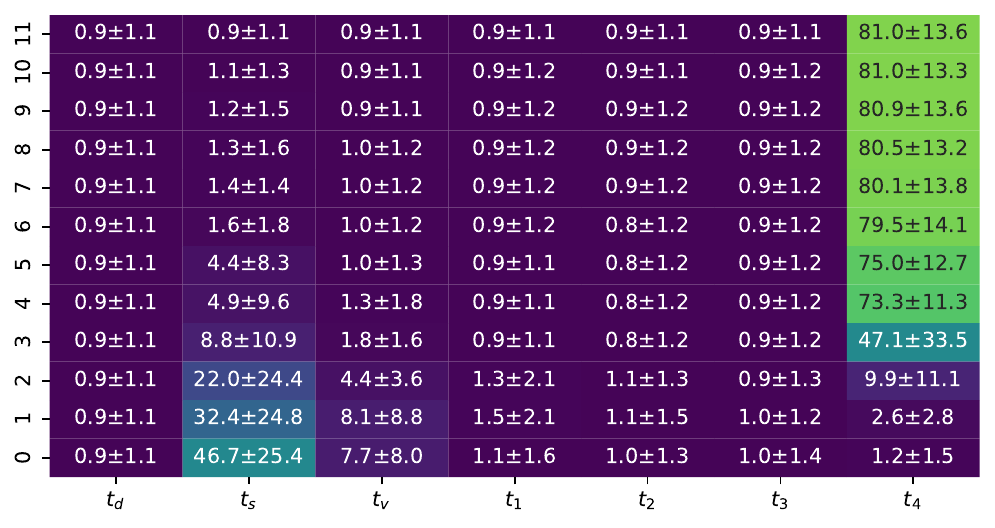}
         \caption{1500 Training Steps.}
     \end{subfigure}
     \begin{subfigure}{0.49\textwidth}
         \centering
         \includegraphics[width=1\textwidth]{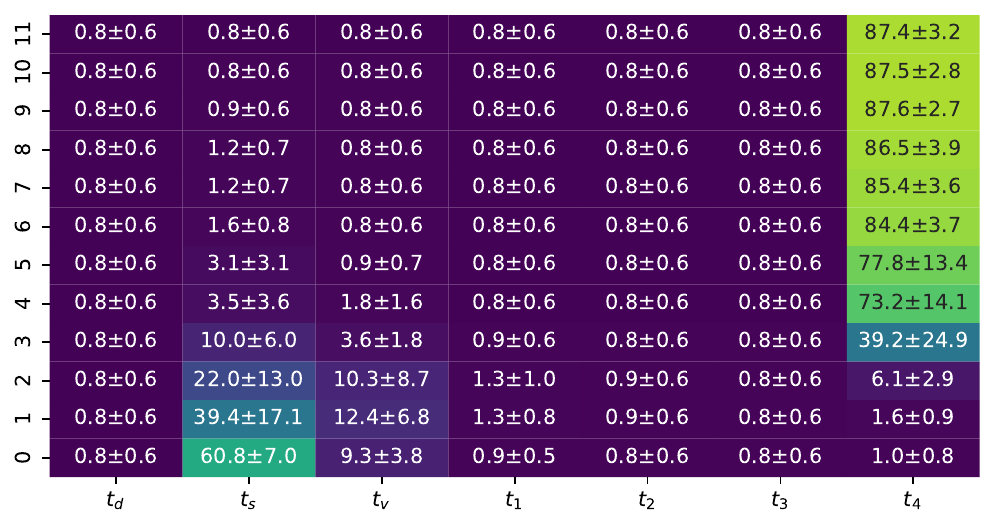}
         \caption{3000 Training Steps.}
     \end{subfigure}
    \caption{Subject--verb agreement interchange intervention accuracies (IIA) for
    \textsc{TokenHop}, with confidence intervals across models trained on 5 different random seeds. Vertical axes denote the GPT-2 layer of the intervention, and horizontal axes denote the token position of the intervention. $t_d$, $t_s$, and $t_v$ represent the tokens for the determiner, subject, and verb. $t_1 \dots t_4$ represent the four tokens/words between the verb.}
    \label{fig:iia_ci_hop_tokens4}
\end{figure*}

\begin{figure*}
    \centering
     \begin{subfigure}{0.49\textwidth}
         \centering
         \includegraphics[width=1\textwidth]{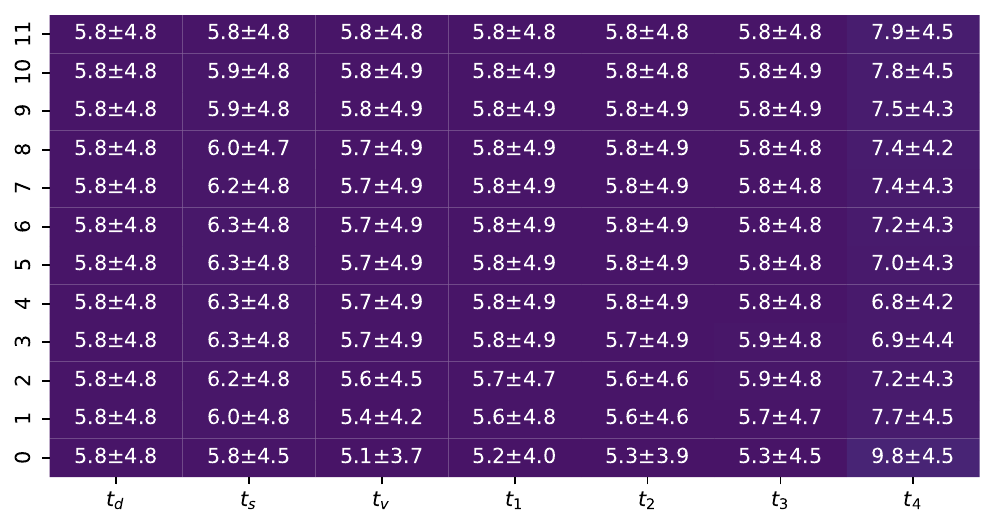}
         \caption{300 Training Steps.}
     \end{subfigure}
     \begin{subfigure}{0.49\textwidth}
         \centering
         \includegraphics[width=1\textwidth]{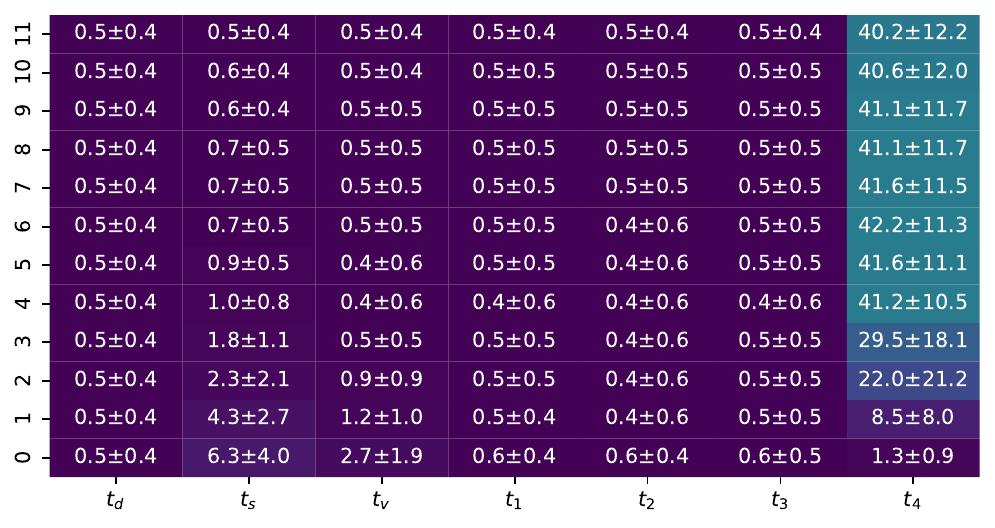}
         \caption{600 Training Steps.}
     \end{subfigure}
     \begin{subfigure}{0.49\textwidth}
         \centering
         \includegraphics[width=1\textwidth]{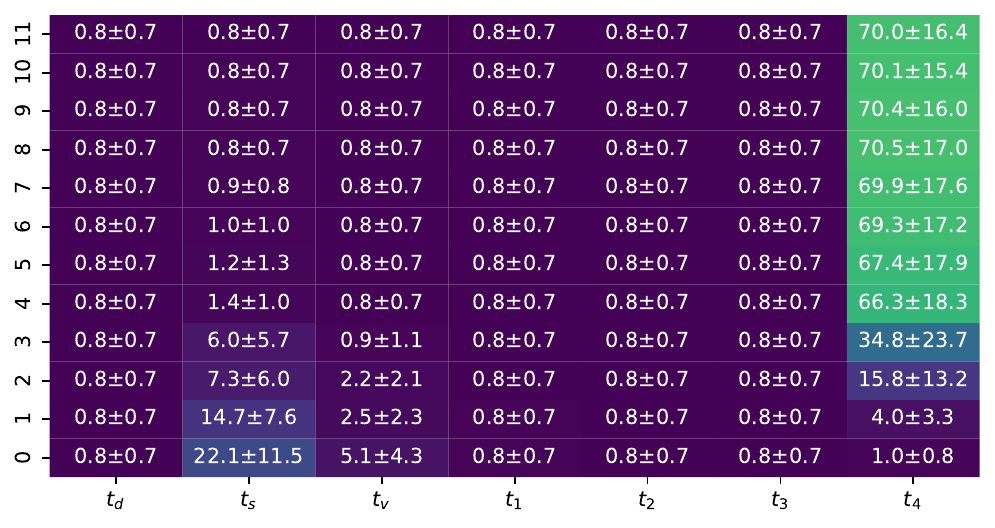}
         \caption{900 Training Steps.}
     \end{subfigure}
     \begin{subfigure}{0.49\textwidth}
         \centering
         \includegraphics[width=1\textwidth]{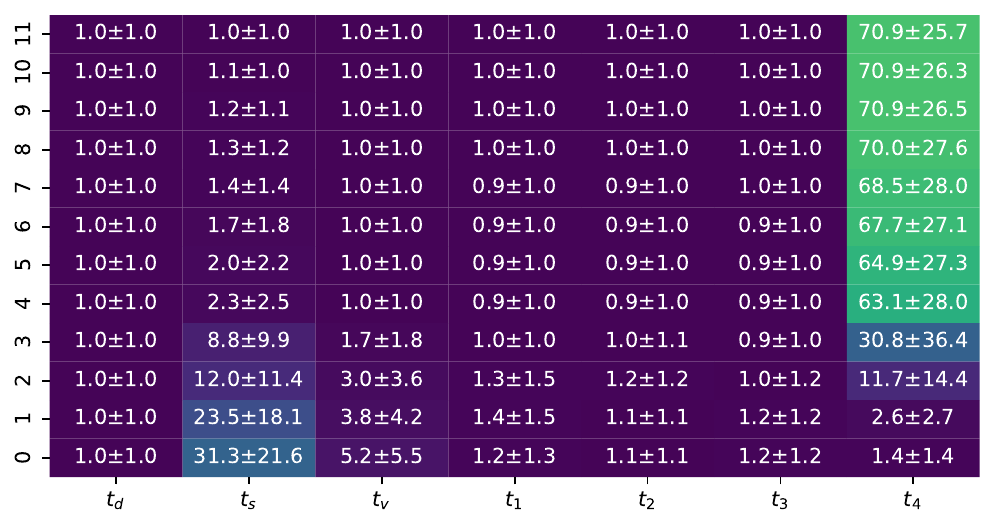}
         \caption{1200 Training Steps.}
     \end{subfigure}
     \begin{subfigure}{0.49\textwidth}
         \centering
         \includegraphics[width=1\textwidth]{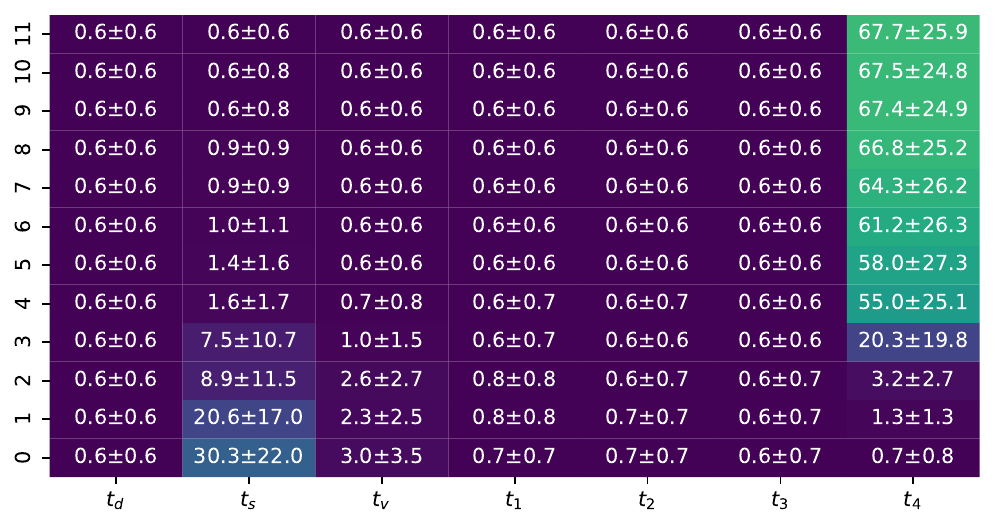}
         \caption{1500 Training Steps.}
     \end{subfigure}
     \begin{subfigure}{0.49\textwidth}
         \centering
         \includegraphics[width=1\textwidth]{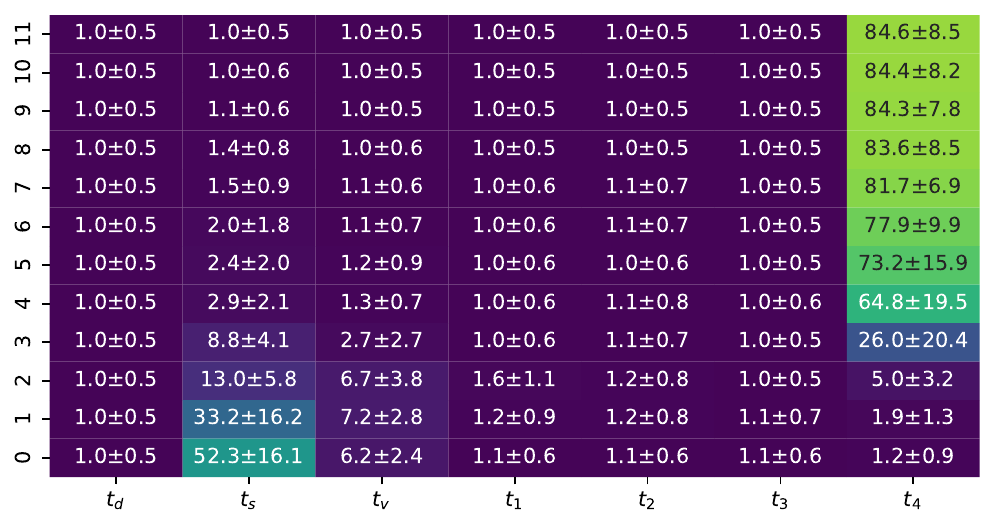}
         \caption{3000 Training Steps.}
     \end{subfigure}
    \caption{Subject--verb agreement interchange intervention accuracies (IIA) for
    \textsc{WordHop}, with confidence intervals across models trained on 5 different random seeds. Vertical axes denote the GPT-2 layer of the intervention, and horizontal axes denote the token position of the intervention. $t_d$, $t_s$, and $t_v$ represent the tokens for the determiner, subject, and verb. $t_1 \dots t_4$ represent the four tokens/words between the verb.}
    \label{fig:iia_ci_hop_words4}
\end{figure*}

\begin{figure*}
    \centering
     \begin{subfigure}{0.3\textwidth}
         \centering
         \includegraphics[width=1\textwidth]{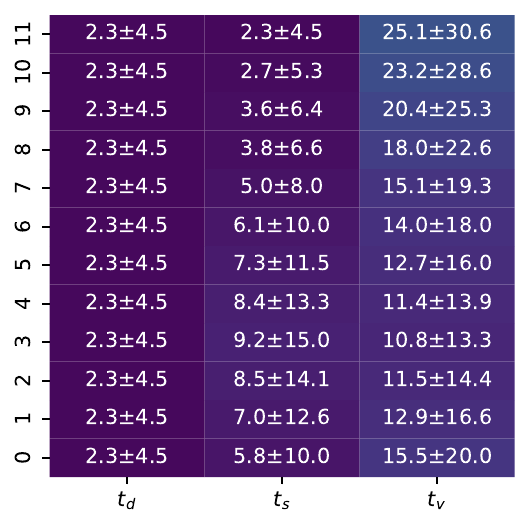}
         \caption{300 Training Steps.}
     \end{subfigure}
     \begin{subfigure}{0.3\textwidth}
         \centering
         \includegraphics[width=1\textwidth]{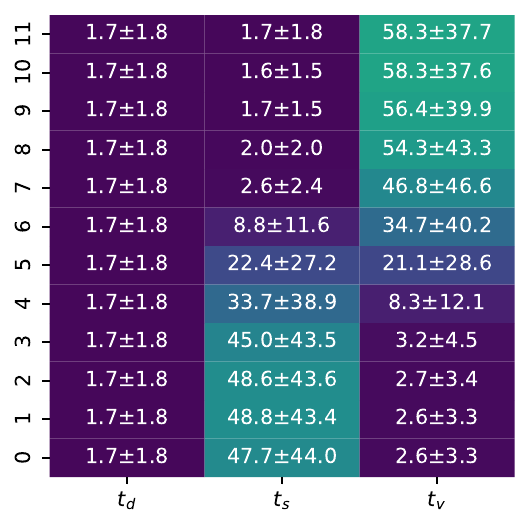}
         \caption{600 Training Steps.}
     \end{subfigure}
     \begin{subfigure}{0.3\textwidth}
         \centering
         \includegraphics[width=1\textwidth]{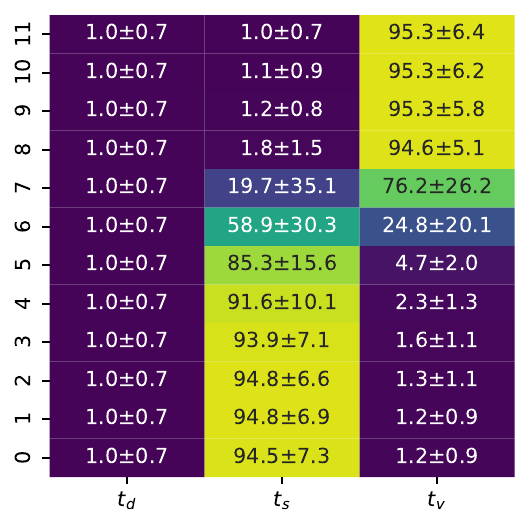}
         \caption{900 Training Steps.}
     \end{subfigure}
     \begin{subfigure}{0.3\textwidth}
         \centering
         \includegraphics[width=1\textwidth]{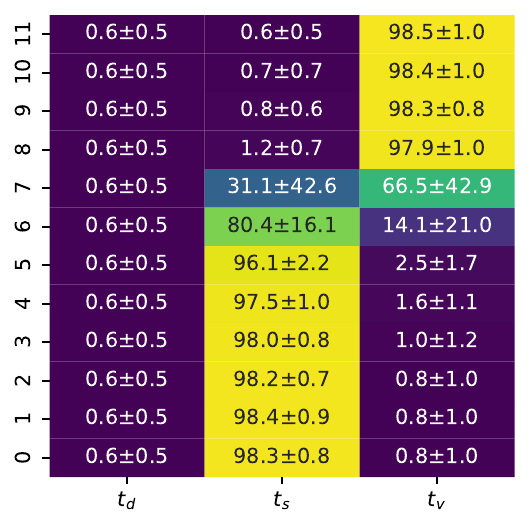}
         \caption{1200 Training Steps.}
     \end{subfigure}
     \begin{subfigure}{0.3\textwidth}
         \centering
         \includegraphics[width=1\textwidth]{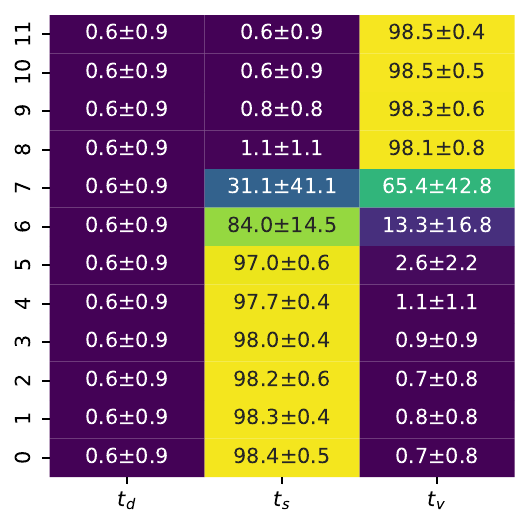}
         \caption{1500 Training Steps.}
     \end{subfigure}
     \begin{subfigure}{0.3\textwidth}
         \centering
         \includegraphics[width=1\textwidth]{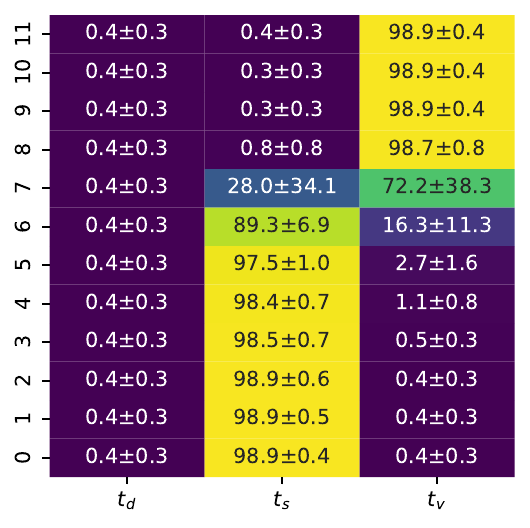}
         \caption{3000 Training Steps.}
     \end{subfigure}
    \caption{Subject--verb agreement interchange intervention accuracies (IIA) for the \textsc{NoHop} model trained \emph{without positional encodings}, with confidence intervals across models trained on 5 different random seeds. Vertical axes denote the GPT-2 layer of the intervention, and horizontal axes denote the token position of the intervention. $t_d$, $t_s$, and $t_v$ represent the tokens for the determiner, subject, and verb, respectively.}
    \label{fig:iia_ci_hop_control_no_pos_encodings}
\end{figure*}

\begin{figure*}
    \centering
     \begin{subfigure}{0.49\textwidth}
         \centering
         \includegraphics[width=1\textwidth]{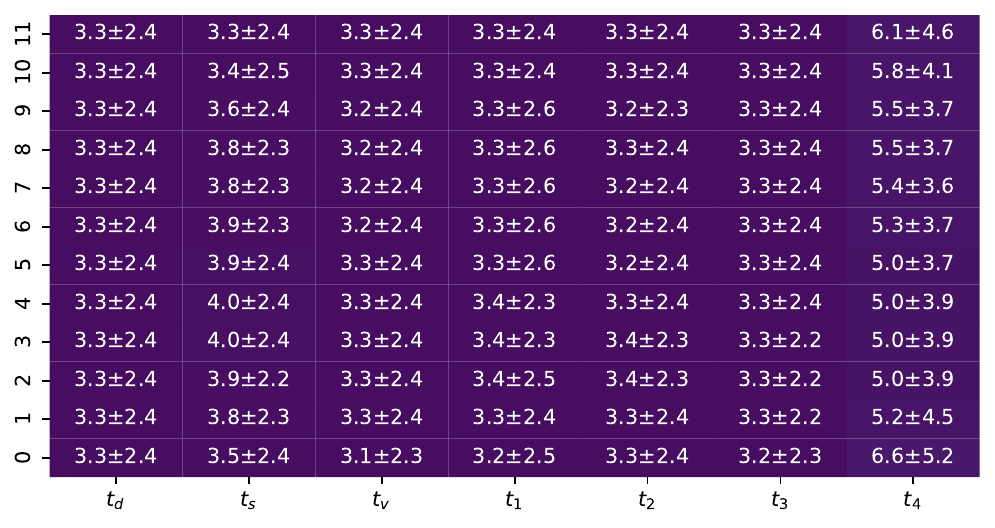}
         \caption{300 Training Steps.}
     \end{subfigure}
     \begin{subfigure}{0.49\textwidth}
         \centering
         \includegraphics[width=1\textwidth]{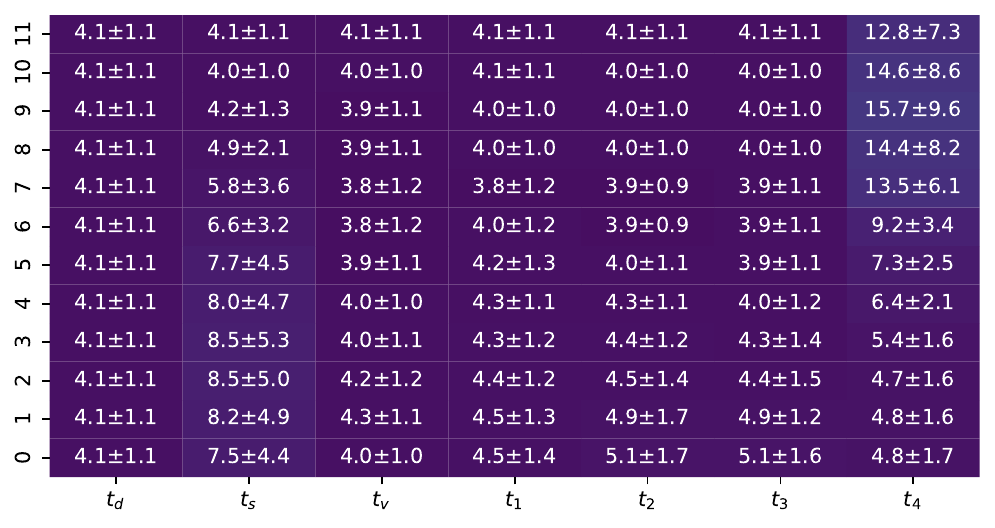}
         \caption{600 Training Steps.}
     \end{subfigure}
     \begin{subfigure}{0.49\textwidth}
         \centering
         \includegraphics[width=1\textwidth]{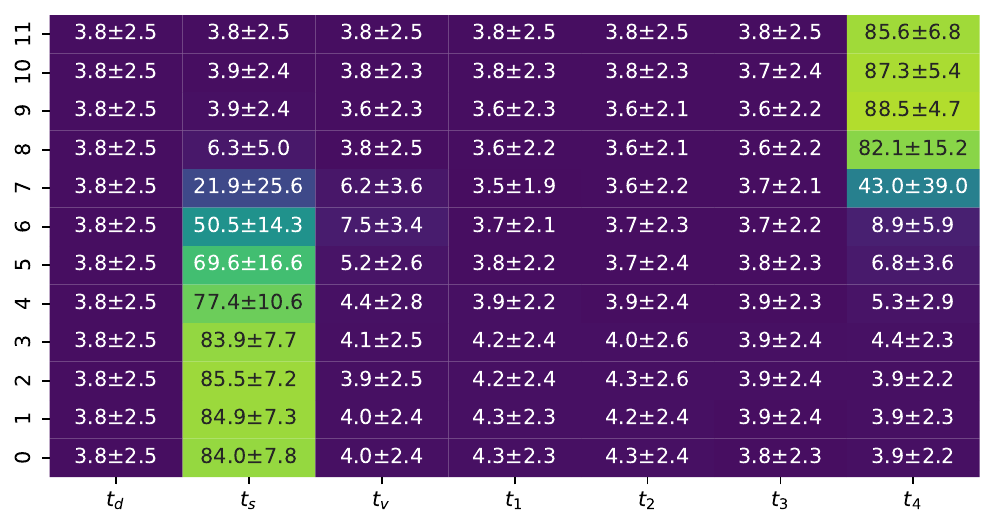}
         \caption{900 Training Steps.}
     \end{subfigure}
     \begin{subfigure}{0.49\textwidth}
         \centering
         \includegraphics[width=1\textwidth]{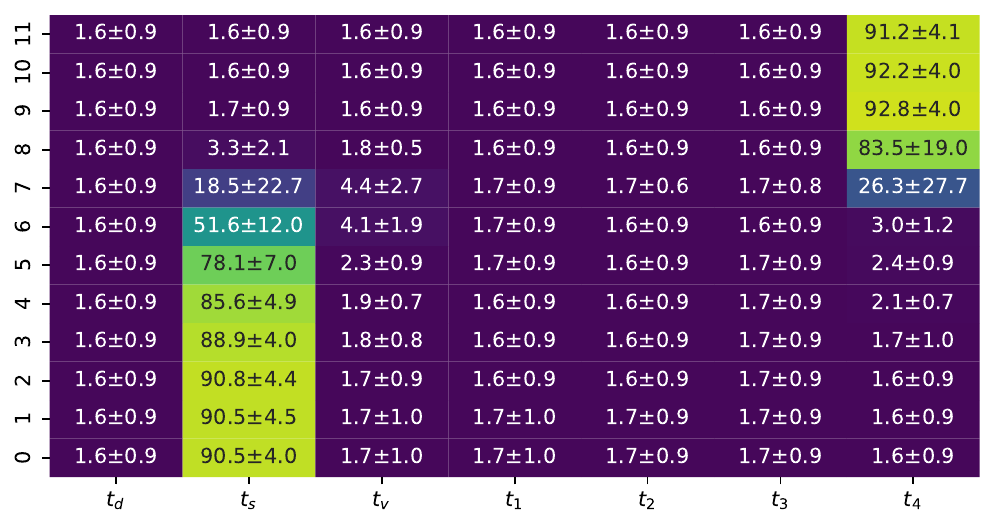}
         \caption{1200 Training Steps.}
     \end{subfigure}
     \begin{subfigure}{0.49\textwidth}
         \centering
         \includegraphics[width=1\textwidth]{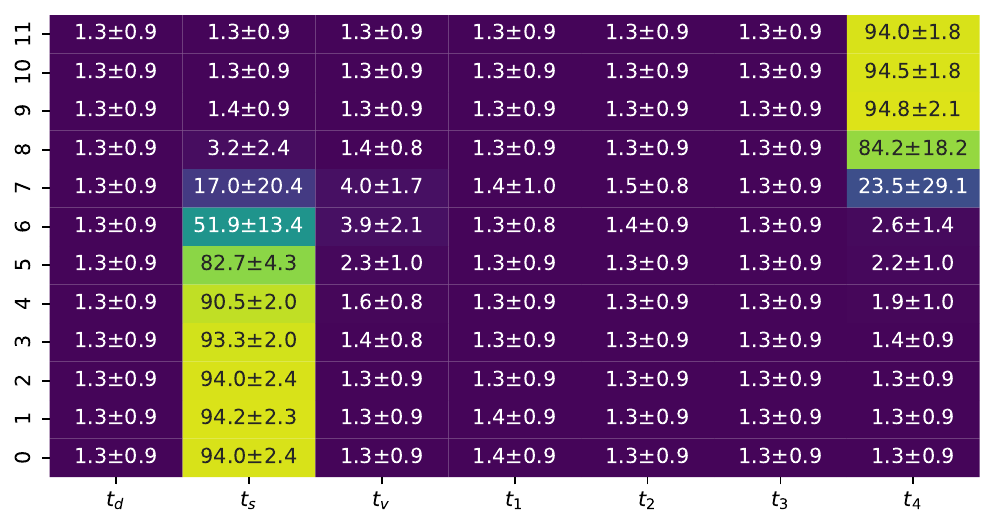}
         \caption{1500 Training Steps.}
     \end{subfigure}
     \begin{subfigure}{0.49\textwidth}
         \centering
         \includegraphics[width=1\textwidth]{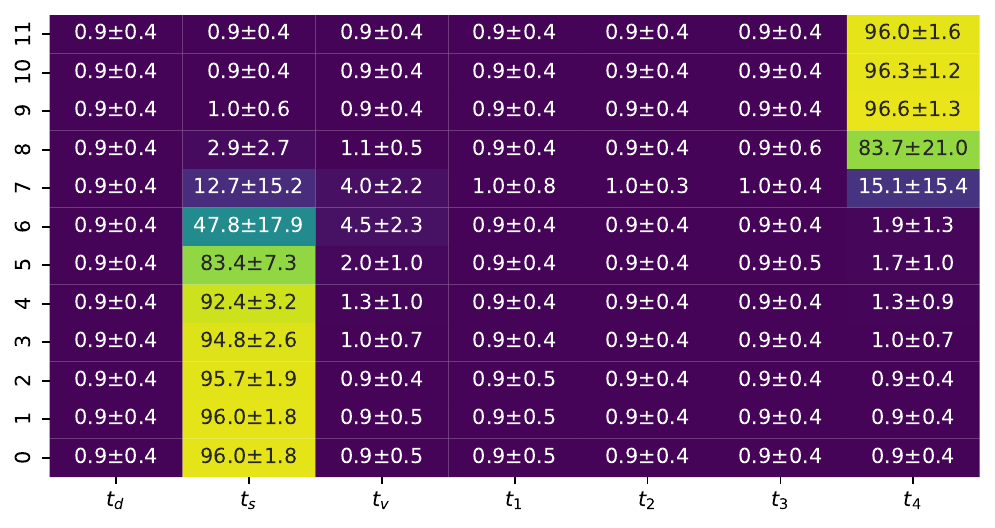}
         \caption{3000 Training Steps.}
     \end{subfigure}
    \caption{Subject--verb agreement interchange intervention accuracies (IIA) for the \textsc{TokenHop} model trained \emph{without positional encodings}, with confidence intervals across models trained on 5 different random seeds. Vertical axes denote the GPT-2 layer of the intervention, and horizontal axes denote the token position of the intervention. $t_d$, $t_s$, and $t_v$ represent the tokens for the determiner, subject, and verb. $t_1 \dots t_4$ represent the four tokens/words between the verb.}
    \label{fig:iia_ci_hop_tokens4_no_pos_encodings}
\end{figure*}

\begin{figure*}
    \centering
     \begin{subfigure}{0.49\textwidth}
         \centering
         \includegraphics[width=1\textwidth]{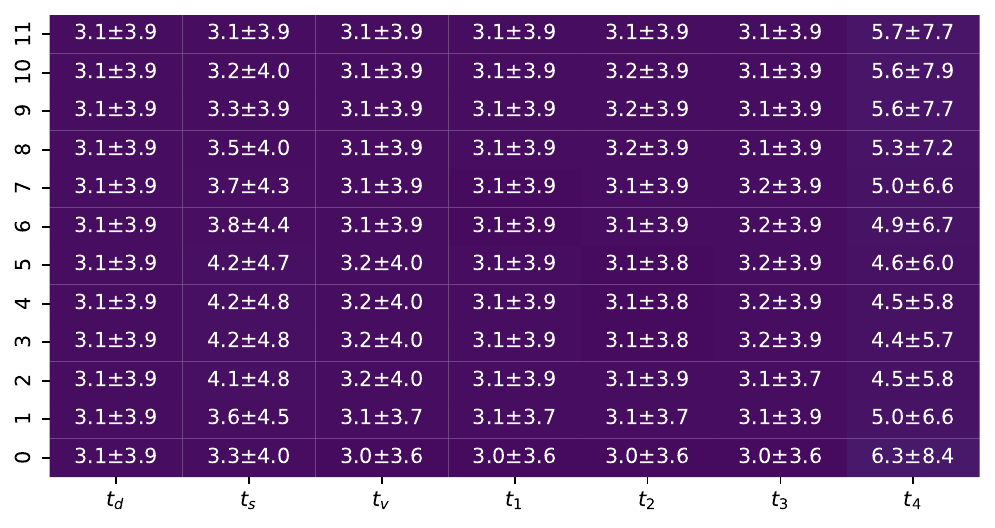}
         \caption{300 Training Steps.}
     \end{subfigure}
     \begin{subfigure}{0.49\textwidth}
         \centering
         \includegraphics[width=1\textwidth]{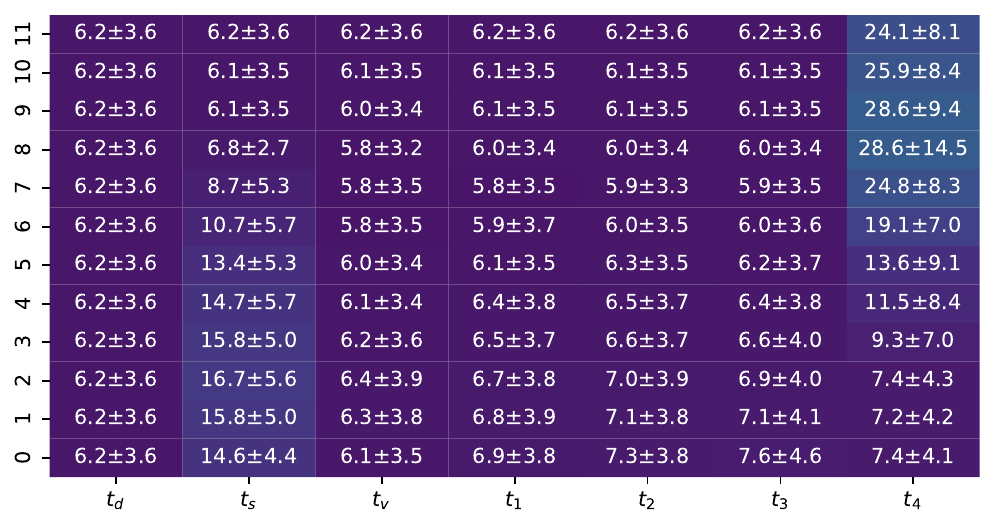}
         \caption{600 Training Steps.}
     \end{subfigure}
     \begin{subfigure}{0.49\textwidth}
         \centering
         \includegraphics[width=1\textwidth]{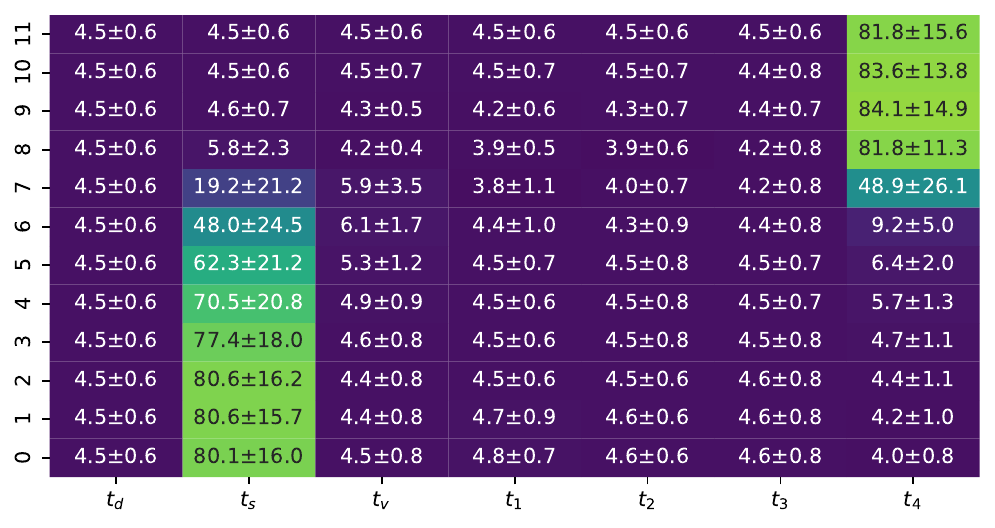}
         \caption{900 Training Steps.}
     \end{subfigure}
     \begin{subfigure}{0.49\textwidth}
         \centering
         \includegraphics[width=1\textwidth]{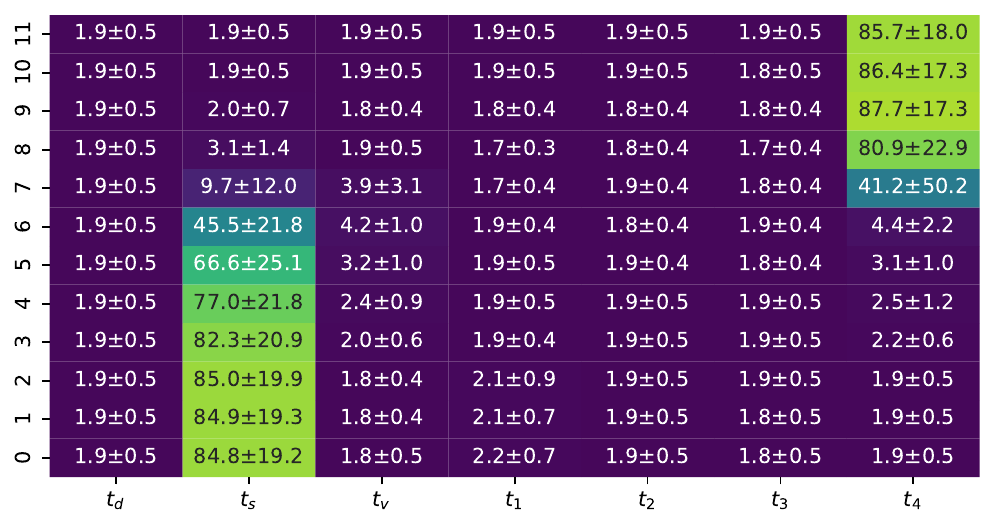}
         \caption{1200 Training Steps.}
     \end{subfigure}
     \begin{subfigure}{0.49\textwidth}
         \centering
         \includegraphics[width=1\textwidth]{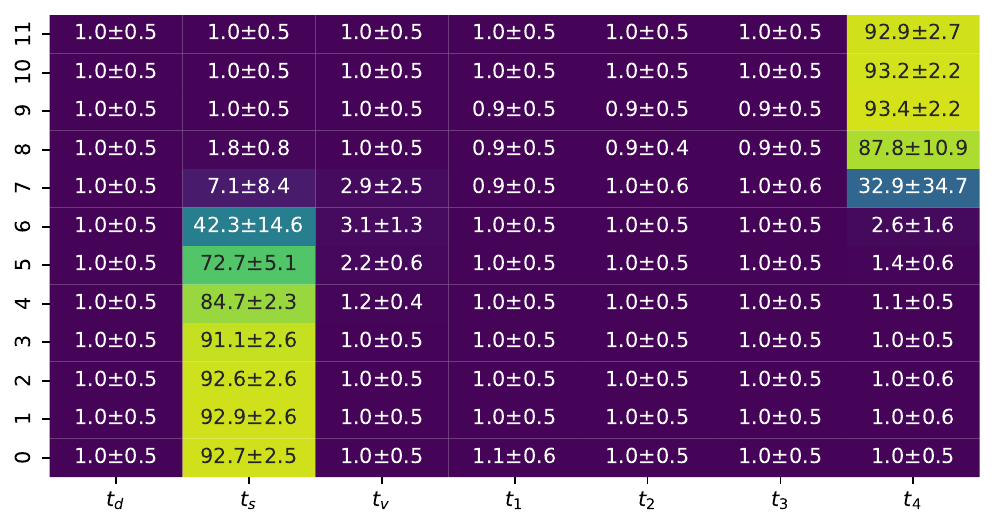}
         \caption{1500 Training Steps.}
     \end{subfigure}
     \begin{subfigure}{0.49\textwidth}
         \centering
         \includegraphics[width=1\textwidth]{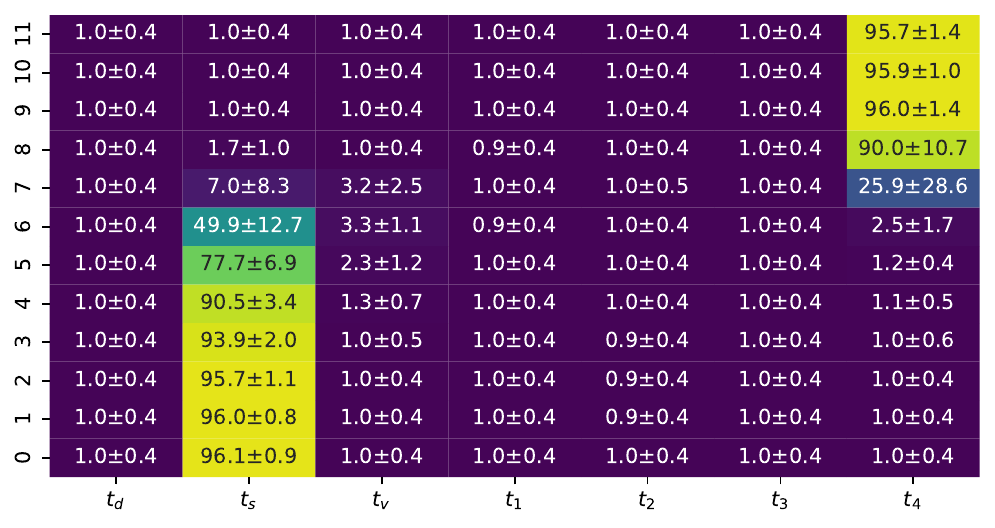}
         \caption{3000 Training Steps.}
     \end{subfigure}
    \caption{Subject--verb agreement interchange intervention accuracies (IIA) for the \textsc{WordHop} model trained \emph{without positional encodings}, with confidence intervals across models trained on 5 different random seeds. Vertical axes denote the GPT-2 layer of the intervention, and horizontal axes denote the token position of the intervention. $t_d$, $t_s$, and $t_v$ represent the tokens for the determiner, subject, and verb. $t_1 \dots t_4$ represent the four tokens/words between the verb.}
    \label{fig:iia_ci_hop_words4_no_pos_encodings}
\end{figure*}

We present the same results of our causal abstraction experiments from \cref{sec:causal-abstractions}, but include confidence intervals for results across models initialized on different random seeds. \Cref{fig:iia_ci_hop_control} presents the results for \textsc{NoHop}; \Cref{fig:iia_ci_hop_tokens4} presents the results for \textsc{TokenHop}; and \Cref{fig:iia_ci_hop_words4} presents the results for \textsc{WordHop}. Figures~\ref{fig:iia_ci_hop_control_no_pos_encodings},~\ref{fig:iia_ci_hop_tokens4_no_pos_encodings}, and~\ref{fig:iia_ci_hop_words4_no_pos_encodings} show the same plots for each \textsc{*Hop} model trained without positional encodings, respectively.

\end{document}